\def\year{2021}\relax
\documentclass[letterpaper]{article} 
\usepackage{aaai21pp}  
\usepackage{times}  
\usepackage{helvet} 
\usepackage{courier}  
\usepackage[hyphens]{url}  
\usepackage{graphicx} 
\usepackage{natbib}  
\usepackage{caption} 
\usepackage{researchpack}
\usepackage{xcolor}
\usepackage{xspace}
\usepackage{hyperref}
\usepackage{subfig}
\usepackage[super]{nth}
\usepackage{wrapfig}
\usepackage{booktabs}

\newcommand{\teodora}[1]{}
\newcommand{\mohit}[1]{}
\newcommand{\stefano}[1]{}

\newcommand{\method}{\textsc{hinter}\xspace}

\title{Machine Guides, Human Supervises:\\ Interactive Learning with Global Explanations}
\author{
    Teodora Popordanoska\textsuperscript{\rm 1},
    Mohit Kumar,\textsuperscript{\rm 1}
    Stefano Teso\textsuperscript{\rm 2}
    \\
}
\affiliations {
    \textsuperscript{\rm 1} KU Leuven \\
    \textsuperscript{\rm 2} University of Trento \\
    \{teodora.popordanoska, mohit.kumar\}@kuleuven.be,
    stefano.teso@unitn.it
}

\begin{document}
\maketitle

\begin{abstract}

    We introduce explanatory guided learning (XGL), a novel interactive learning strategy in which a machine guides a human supervisor toward selecting informative examples for a classifier.
    The guidance is provided by means of global explanations, which summarize the classifier's behavior on different regions of the instance space and expose its flaws.
    Compared to other explanatory interactive learning strategies, which are machine-initiated and rely on local explanations, XGL is designed to be robust against cases in which the explanations supplied by the machine oversell the classifier's quality.
    Moreover, XGL leverages global explanations to open up the black-box of human-initiated interaction, enabling supervisors to select informative examples that challenge the learned model.
    By drawing a link to interactive machine teaching, we show theoretically that global explanations are a viable approach for guiding supervisors.
    Our simulations show that explanatory guided learning avoids overselling the model's quality and performs comparably or better than machine- and human-initiated interactive learning strategies in terms of model quality.

\end{abstract}

\section{Introduction}

The increasing ubiquity and sophistication of machine learning calls for strategies to understand and control predictors learned from data.  Recent work has shown the promise of combining \emph{local explanations} with \emph{interactive learning}~\cite{teso2019explanatory,schramowski2020right}.  Existing implementations of this idea extend active learning by enabling the machine to not only choose its queries, but also present predictions for the queries and explanations for those predictions to the user.  The explanations continuously illustrate the classifier's beliefs relative to the queries.  In addition, the user supplies feedback on both predictions and explanations, driving the model away from bad hypotheses.

This works very well in some scenarios~\cite{schramowski2020right}, but it becomes problematic when the model is affected by unknown unknowns, that is, (regions of) high-confidence mistakes~\cite{dietterich2017steps}.  In this case the machine tends to only query about those mistakes that it is aware of~\cite{attenberg2010label}.  Since the local explanations focus on the queries, the ``narrative'' output by the machine mostly ignores the unknown unknowns and can thus inadvertently over-sell the performance of the classifier.  We call this phenomenon \emph{narrative bias} (NB).

As a remedy, we introduce \emph{explanatory guided learning} (XGL), an interactive learning protocol in which the supervisor is responsible for choosing the examples and the machine guides the supervisor using \emph{global explanations} that summarize the whole decision surface of the predictor.  XGL brings two key benefits.  First, since global explanations do not focus on individual queries, they are not affected by NB;  this helps users to build less biased expectations of the classifiers' behavior even in the presence of unknown unknowns.  Second, XGL extends human-initiated interactive learning with explanations, thus helping supervisors to identify mistakes made by the model.  A theoretical analysis based on a link to interactive machine learning validates the usefulness of global explanations for designing high-quality training sets and highlights a natural trade-off between complexity of the explanations and quality of the supervision.
Our experiments on synthetic and real-world data show that XGL helps (even noisy) simulated users to select informative examples even in the presence of unknown unknowns.

Summarizing, we:
1) Identify the issue of narrative bias;
2) Introduce explanatory guided learning, which combines human-initiated interactive learning with machine guidance in the form of global explanations;
3) Provide a theoretical analysis that highlights the viability of global explanations for designing high-quality training sets;
4) Compare \method, a rule-based implementation of XGL, against human- and machine-initiated interactive learners on a synthetic and several real-world data sets.

\section{Interaction, Explanations, and the Unknown}

\begin{figure*}[t]
    \centering
    \begin{minipage}{0.25\textwidth}
        \centering
        \includegraphics[width=\linewidth]{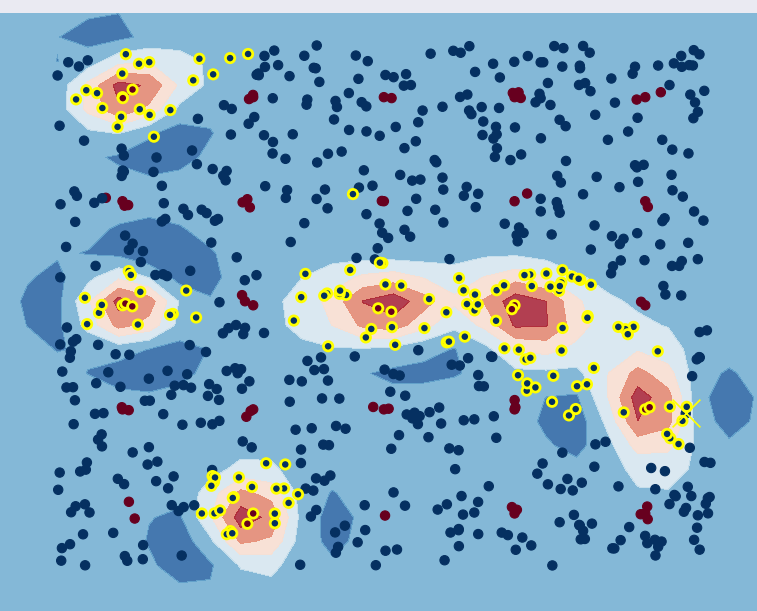}
    \end{minipage}
    \begin{minipage}{0.25\textwidth}
        \centering
        \includegraphics[width=\linewidth]{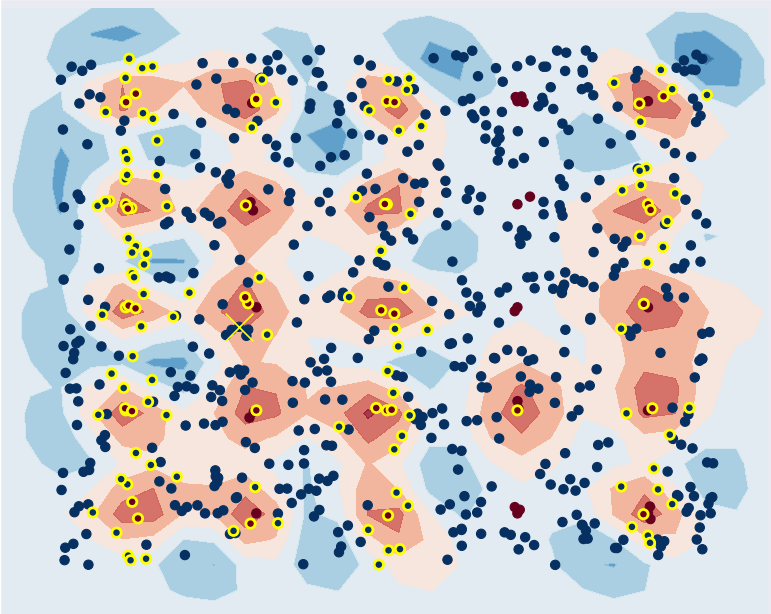}
    \end{minipage}
    \begin{minipage}{0.25\textwidth}
        \centering
        \includegraphics[width=2.5\linewidth]{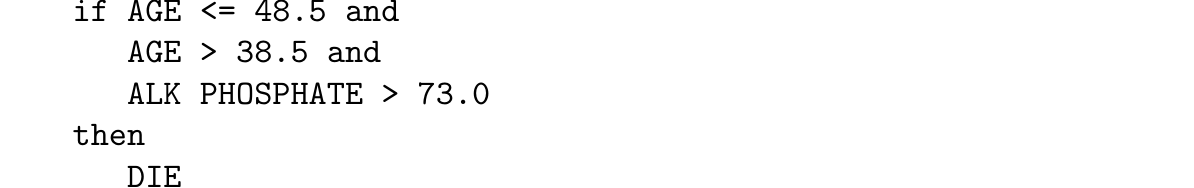}
        \vspace{3em}
    \end{minipage}
    \caption{Left:  uncertainty-based AL queries points (circled in yellow) around known red clusters and ignores the unknown ones, even after 140 iterations.  Middle:  XGL discovers most red clusters.  Right:  example rule extracted by \method from the hepatitis data set (classes are LIVE, DIE):  it takes little effort for a medical doctor to understand and (in)validate such a rule.}
    \label{fig:example}
\end{figure*}

Let $\calH$ be a class of black-box classifiers $h: \bbR^d \to \{0, 1\}$, e.g., neural nets or kernel machines.  Our goal is to learn a classifier $h \in \calH$ from data.  Initially only a small training set $S_0 = \{ (\vx_i, y_i) \}_{i=1}^{n_0}$ is available, but more examples can be requested from a supervisor.
In order to facilitate understanding and control, the machine is additionally required to explain its own beliefs in a way that is
understandable to an expert supervisor and
useful for identifying mistakes in the logic of the predictor.

These requirements immediately rule out strategies like active learning (AL)~\cite{settles2012active,hanneke2014theory} and guided learning (GL)~\cite{attenberg2010label}, in which the model is treated as a black-box.
A better fit is explanatory active learning (XAL)~\cite{teso2019explanatory,schramowski2020right}.  As in AL, in XAL the machine iteratively selects queries $\vx$ from a pool of unlabeled instances and asks the supervisor to label them, but in addition it also supplies \emph{predictions} for the queries and \emph{local explanations}~\cite{guidotti2018survey} for the predictions.
The explanations expose the reasons behind individual predictions in terms of interpretable factors, like feature relevance~\cite{ribeiro2016should}, and together with the predictions establish a \emph{narrative} that enables the supervisor to build expectations about the classifier.
Moreover, the supervisor can control the predictor by providing feedback on the explanations, for instance by indicating what features the predictor is wrongly relying on (confounders).

\paragraph{Narrative Bias:}  AL struggles in the presence of unknown unknowns (UUs), that is, regions in which the classifier makes high-confidence mistakes~\cite{dietterich2017steps}.  These are common in the presence of class skew~\cite{attenberg2010label} and concept drift~\cite{gama2014survey,boult2019learning} and are especially challenging when associated with a high mislabeling cost.
The reason is that classifiers affected by UUs are fundamentally unaware of their own faults, and therefore cannot intentionally select queries that expose these mistakes~\cite{attenberg2010label,attenberg2015beat}.

A so-far unexplored consequence of this phenomenon is that, since the local explanations focus on the queries, the ``narrative'' output by XAL ignores the UUs, where by definition the machine performs poorly.  Hence, UUs may induce the machine to unwillingly oversell its own performance to the user, especially if they are associated with a high cost.
This leads to narrative bias (NB).  Intuitively, NB measures the difference between the performance conveyed by the queries $\vx_1, \ldots, \vx_T$ to the user and the true risk $R_T = \bbE_{\vx \sim P}[L_T(\vx)]$, where $P$ is the ground-truth distribution and $L_t(\cdot)$ is the loss incurred by the classifier learned at iteration $t = 1, \ldots, T$.  Arguably, the performance perceived by the user is a function of the losses $\{L_t(\vx_t)\}_{t=1}^T$ exposed by the narrative of XAL over time.\footnote{While we will focus on labels-induced NB, for simplicity, our arguments apply to NB induced by local explanations, which can be defined analogously.  This is however harder to assess empirically since most datasets do not include ground-truth explanations.}  Despite some degree of subjectivity, it is reasonable to define NB as $\frac{1}{T} \sum_{t=1}^T L_t(\vx_t) - R_T$, although alternatives can be conceived.

Figure~\ref{fig:example} (left) illustrates this issue on synthetic data designed to induce unknown unknowns.  The red examples are grouped in evenly spaced clusters while the blue examples are distributed uniformly everywhere else.  The queries chosen by an active RBF SVM after 140 iterations of uncertainty sampling are circled in yellow and the decision surface is shown in the background.  The queries clearly concentrate around the \emph{known} red clusters, where the classifier already performs well in terms of both predictions and explanations (e.g., feature relevance or gradient information).  The bad performance of the model on the \emph{unknown} red clusters is completely ignored by the queries and hence by the narrative output by XAL.  Notice that the core issue is not uncertainty sampling itself:  indeed, representative sampling strategies like density-weighted AL~\cite{fu2013survey} also struggle with UUs~\cite{attenberg2010label} and can therefore be affected by narrative bias.


\section{Explanatory Guided Learning}

We propose to use  \emph{human-initiated} interactive learning as an antidote to narrative bias.  The intuition is straightforward:  if a motivated and knowledgeable supervisor could see and understand the decision surface of $h$, she could recognize both known and unknown mistakes -- and hence determine whether the predictor misbehaves -- and intelligently select examples that correct them.  Of course, since the decision surface of $h$ can be very complex, this strategy is purely ideal.  The challenge is then how to make it practical.

\paragraph{Global Explanations:}  We propose to solve this issue by summarizing $h$ in a compact and interpretable manner using \emph{global explanations}~\cite{andrews1995survey,guidotti2018survey}.  A global explanation is an interpretable surrogate $g \in \calG$ of $h$, usually a shallow decision tree~\cite{craven1996extracting,krishnan1999extracting,boz2002extracting,tan2016tree,bastani2017interpreting,yang2018global} or a rule set~\cite{nunez2002rule,johansson2004accuracy,barakat2010rule,augasta2012reverse}.\footnote{Global explanations based on feature dependencies or shape constraints~\cite{henelius2014peek,tan2018learning} will not be considered.}  What makes these models attractive is that they decompose into simple atomic elements, like short decision paths or simple rules, that can be described and visualized independently and associated to individual instances.  Figure~\ref{fig:example} (right) illustrates an example rule.
Usually $g$ is obtained via model distillation~\cite{bucilua2006model}, that is, by projecting $h$ onto $\calG$ using a global explainer $\pi: \calH \to \calG$, defined as:
\begin{align}
    \textstyle
    \pi(h)
        & \textstyle \defeq \argmin_{g' \in \calG} M(h, g') + \lambda \Omega(g'),
    \\
    M(h, g)
        & \textstyle \defeq \bbE_{\vx \sim P} [M(h(\vx), g(\vx))].
    \label{eq:distillation}
\end{align}
Here $P$ is the ground-truth distribution, $M$ is an appropriate loss function, $\Omega(\cdot)$ measures the complexity of the explanation, and $\lambda > 0$ controls the trade-off between faithfulness to $h$ and simplicity.  The expectation is typically replaced by an empirical Monte Carlo estimate using fresh i.i.d. samples from $P$ or using any available unlabeled instances.

\paragraph{The Algorithm:}
The pseudo-code of XGL is listed in Algorithm~\ref{alg:xgl}.  In each iteration, a classifier $h$ is fit on the current training set $S$ and summarized using a global explanation $g = \pi(h)$.  Then, $g$ is presented to the supervisor.  Each rule is translated to a visual artifact or to a textual description and shown together with the instances it covers.  The instances are labeled in accordance to the \emph{rule}.  The supervisor is then asked for one or more examples on which the explanation is mistaken,  which are added to the training set $S$.  The loop repeats until $h$ is good enough or the query budget is exhausted.

\begin{algorithm}[h]

    \begin{algorithmic}[1]
        \State $S \gets S_0$ \label{line:initialize}
        \Repeat
            \State fit $h$ on $S$ \label{line:updatef}
            \State compute $g = \pi(h)$ \Comment{Eq.~\eqref{eq:distillation}} \label{line:updateg}
            \State present $g$ to the supervisor
            \State receive (possibly high-confidence) mistakes $S'$ \label{line:query}
            \State $S \gets S \cup S'$ \label{line:receive}
        \Until{query budget exhausted or $h$ good enough}
    \end{algorithmic}

    \caption{\label{alg:xgl}  Explanatory guided learning.  $S_0$ is the initial training set and $\pi$ is a global explainer.}

\end{algorithm}

\noindent In practice, the supervisor can search for mistakes by:
\begin{itemize}

    \item \emph{scanning through the instances}, each shown together with a prediction and a rule, and pointing out one or more mistakes, or
    
    \item \emph{searching for a wrong rule} and then supplying a counter-example for it.
    
\end{itemize}
The first strategy mimics guided learning (GL)~\cite{attenberg2010label}:  in GL, given a textual description of some target concept and a list of instances obtained using a search engine, the user has to recognize instances of that concept in the list.  The difference is that in XGL the instances are presented together with a corresponding prediction and explanation, which makes it possible for the user to identify actual mistakes -- which in GL is not possible -- and to gain insight into the model.  In this sense, XGL is to GL what XAL is to AL:  an approach for making the interaction less opaque.  Instances can be grouped by rule to facilitate scanning through them.  Given that GL was successfully deployed in industrial applications~\cite{attenberg2010label}, arguably XGL also can be.

The second strategy is geared toward experts capable of recognizing bad rules and identifying or synthesizing counter-examples.  Since there usually are far fewer rules than instances (in our experiments, usually 5-30 rules versus hundreds or thousands of instances), this can be more efficient, at least as long as interpreting individual rules is not too taxing.  Interpretability can be facilitated by regularizing the rules appropriately.

\paragraph{Advantages and Limitations:}
XGL is designed to be robust against narrative bias while enabling expert supervisors to identify mistakes.  We stress that simply combining global explanations with machine-initiated interactive learning would \emph{not} achieve the same effect, as the choice of queries would still be affected by UUs.
Another benefit of XGL is that it natively supports selecting \emph{batches} of instances 
in each iteration, thus amortizing the cost of queries.  Indeed, this is the most natural usage of XGL.  Nevertheless, we restrict our discussion and experiments to the one-example-per-query case to simplify the comparison with the competitors.

Shifting the responsibility of choosing the examples to a human supervisor is not devoid of risks.  A global explanation might be too rough a summary or it may be misunderstood by the supervisor.  These issues, however, affect AL and XAL too:  the annotator's performance can be poor even in black-box interaction~\cite{zeni2019fixing} and local explanations can be unfaithful~\cite{teso2019toward,dombrowski2019explanations}.  As with all approaches, XGL should be applied in settings where these issues are unlikely or their effects are tolerable.

The main downside of XGL is undoubtedly the cognitive and computational cost of global explanations.
The computational cost can be reduced by updating $g$ incrementally as $h$ is updated.
The cognitive cost can be improved in several ways.  For instance, the global explanations can be restricted to those regions of the instance space that contain, e.g., high cost instances.
A more flexible alternative is to adapt the resolution of the global explanations on demand:  one could supply coarse rules $g$ to the supervisor, and then allow him or her to refine $g$ and ``zoom into'' those regions or subspaces that appear suspicious, as has been proposed in the context of local~\cite{lee2019understanding,hase2019interpretable} and global explanations~\cite{lakkaraju2017identifying}.

Regardless, global explanations are necessarily more demanding than local explanations or no explanations.  Like other interaction protocols~\cite{lage2018human}, XGL involves an ``effortful'' human-in-the-loop step in which the supervisor must invest time and attention.  Our argument is that this extra effort is justified in applications in which the cost of overestimating a misbehaving model is large.

Ultimately, given their complementary qualities (lower vs higher cognitive overhead, lower vs higher robustness to narrative bias), XGL should not be viewed as an alternative to AL or XAL, but rather as a \emph{supplement} to them.  One option is to interleave XAL and XGL in a mixed-initiative fashion:  the interaction would normally be machine-initiated and switch to XGL either periodically, when the user requests a global explanation, or when the machine realizes that machine-selected supervision has little effect and hence that human-selected examples are required.  This would substantially decrease the cognitive cost of XGL while retaining most of its benefits.  Here, however, we focus on XGL and leave a study of mixed-initiative strategies to future research.

\subsection{Theoretical Analysis}

 Are global explanations useful for identifying good examples?  To answer this question, we draw a connection between XGL and interactive machine teaching.  Machine teaching is the problem of designing an optimal training set (aka \emph{teaching set}) $S(h^*)$ for teaching a target hypothesis $h^* \in \calH$ to a (consistent) learner~\cite{zhu2015machine}.  If the learner is black box, then passive teachers oblivious to the machine's beliefs cannot perform better than random sampling.  On the other hand, interactive teachers that have access to the decision surface of the black-box perform almost optimally~\cite{melo2018interactive,chen2018near,dasgupta2019teaching}.  Since global explanations approximate this decision surface, we expect them to offer similar benefits.

Let $\calX$ be the set of possible instances and $S(h^*) \subseteq \calX$ be a teaching set for $h^* \in \calH$.\footnote{The sets $\calX$, $\calH$ and $\calG$ are assumed to be finite.  The infinite case can be recovered using standard arguments from statistical learning theory, see~\cite{dasgupta2019teaching}.}  We make use of the following key result by~\cite{dasgupta2019teaching}:

\begin{thm}
\label{thm:original}
Let $h^* \in \calH$ be the ground-truth classifier and $\delta \in (0, 1)$.  There exists a teaching oracle that with probability at least $1 - \delta$ halts after at most $|S(h^*)| \lg 2 |\calX|$ iterations and outputs a training set $S$ of expected size at most $(1 + |S(h^*)| \lg 2|\calX|) (\ln |\calH| + \ln \frac{1}{\delta})$ that teaches any consistent learner to perfectly recover $h^*$.
\end{thm}

Hence, with high probability the size of $S$ is at worst a $\ln |\calX| \ln |\calH|$ factor off from that of the teaching set $S(h^*)$.   The teaching oracle mentioned in the theorem builds $S$ by iteratively comparing the decision surface of the current classifier $h$ to that of $h^*$ and then adding any misclassified points to the training set with a carefully designed probability.  The black-box $h$ is then updated using the new training set.  The loop repeats until there are no misclassified points.  See the Supplementary Material for a detailed description.

The case of an interactive teacher with access to global explanations can be reduced to the above by: i)~computing a global explanation for the target concept $g^* = \pi(h^*)$, ii)~running the teaching oracle mentioned in the theorem to produce an almost-optimal training set for $g^*$, and iii)~finding an $h \in \calH$ such that $g = \pi(h)$.  (This becomes trivial if $\calG \subseteq \calH$.)  There is one complication:  now the teaching oracle can only observe and provide feedback relative to the difference between $g$ and $g^*$.  This introduces two irreducible approximations, one between $g$ and $h$ and the other between $g^*$ and $h^*$.  For standard losses (like the $0$-$1$ loss) the triangle inequality implies
$
    L(h, h^*) \le L(h, g) + L(g, g^*) + L(g^*, h^*) \le L(g, g^*) + 2\rho
$,
where $\rho \defeq \max_{h \in \calH} L(h, \pi(h))$.  Therefore, our reduction guarantees a $2\rho$-approximation teaching oracle.  More formally:
 
\begin{proposition}
    \label{thm:new}
    Let $h^* \in \calH$ be the ground-truth classifier, $\pi: \calH \to \calG$ a global explainer, and $\delta \in (0, 1)$.  There exists a teaching oracle that with probability at least $1 - \delta$ halts after at most $|S(g^*)| \lg 2 |\calX|$ iterations and outputs a training set $S$ of expected size at most $(1 + |S(g^*)| \lg 2|\calX|) (\ln |\calG| + \ln \frac{1}{\delta})$ that teaches any consistent learner a hypothesis $h$ that satisfies $L(h, h^*) \le 2\rho$.
\end{proposition}

Since the global explanations act as summaries, $|\calG|$ is presumably smaller than $|\calH|$ (or more generally the Vapnik-Chervonenkis dimension of $\calG$ is smaller than that of $\calH$), which also implies $|S(g^*)| \le |S(h^*)|$.  The proposition validates the use of global explanations for interactive learning:  compared to Theorem~\ref{thm:original}, learning with global explanations requires \emph{fewer} iterations and examples, at the cost introducing an approximation factor, as expected.  At the same time, $\rho$ depends on how good of a summary $\calG$ can be compared to $\calH$.  
Interestingly, $\rho$ could be reduced  by dynamically adapting the resolution of global explanations, as hinted to in the previous Section.

\section{Experiments}

We answer empirically the following research questions:
\textbf{Q1}: Is XGL robust against narrative bias?
\textbf{Q2}: Is XGL competitive with AL and GL in terms of sample complexity and model quality?
\textbf{Q3}: How does the supervisor's performance affect the effectiveness of XGL?

Our rule-based implementation of XGL, named \method (Human-INiTiated Explanatory leaRning), was compared against several human- and machine-initiated alternatives on several UCI data sets using standard binary classifiers (SVMs and gradient boosting trees).  All results are $5$-fold cross-validated using stratification.  For each fold, the training set initially includes five examples, at least two per class.  More training examples were acquired by querying a simulated supervisor, as typically done in interactive learning, one query per iteration for $100$ iterations.  The simulator returned labels or instances from the unlabelled set according to the chosen interaction protocol, see below.  More details about \method, competitors and data sets are left to the Supplementary Material.   The code and experiments can be found at:  GITHUB REPO.

\textbf{Data sets:}  The synthetic data set, illustrated in Figure~\ref{fig:example}, consists of an unbalanced collection of $941$ blue and $100$ red bi-dimensional points, with a class ratio of about $10:1$.  The red points were sampled at random from 25 Gaussian clusters aligned on a five by five grid.  The blue points were sampled uniformly from outside the red clusters.  We also consider several classification data sets from the UCI repository~\cite{Dua:2019} with in-between 150 and 48842 examples and 4 and 30 features.
To keep the run-time manageable, a couple of data sets were sub-sampled to $10\%$ of their original size using stratification.  We are interested in measuring the effect of UUs, but not all data sets induce high-cost UUs in our classifiers.  Hence, we injected in each data set ``disjoint clusters'' -- like in our synthetic data set -- by flipping the label of $10$ random clusters out of $100$ clusters obtained with $k$-means.  The weight of the flipped examples was increased to $25$ to simulate high-cost UUs.  Results for both the original and modified data sets are reported.

\begin{figure*}[tb]
    \centering
    \includegraphics[width=0.8\linewidth]{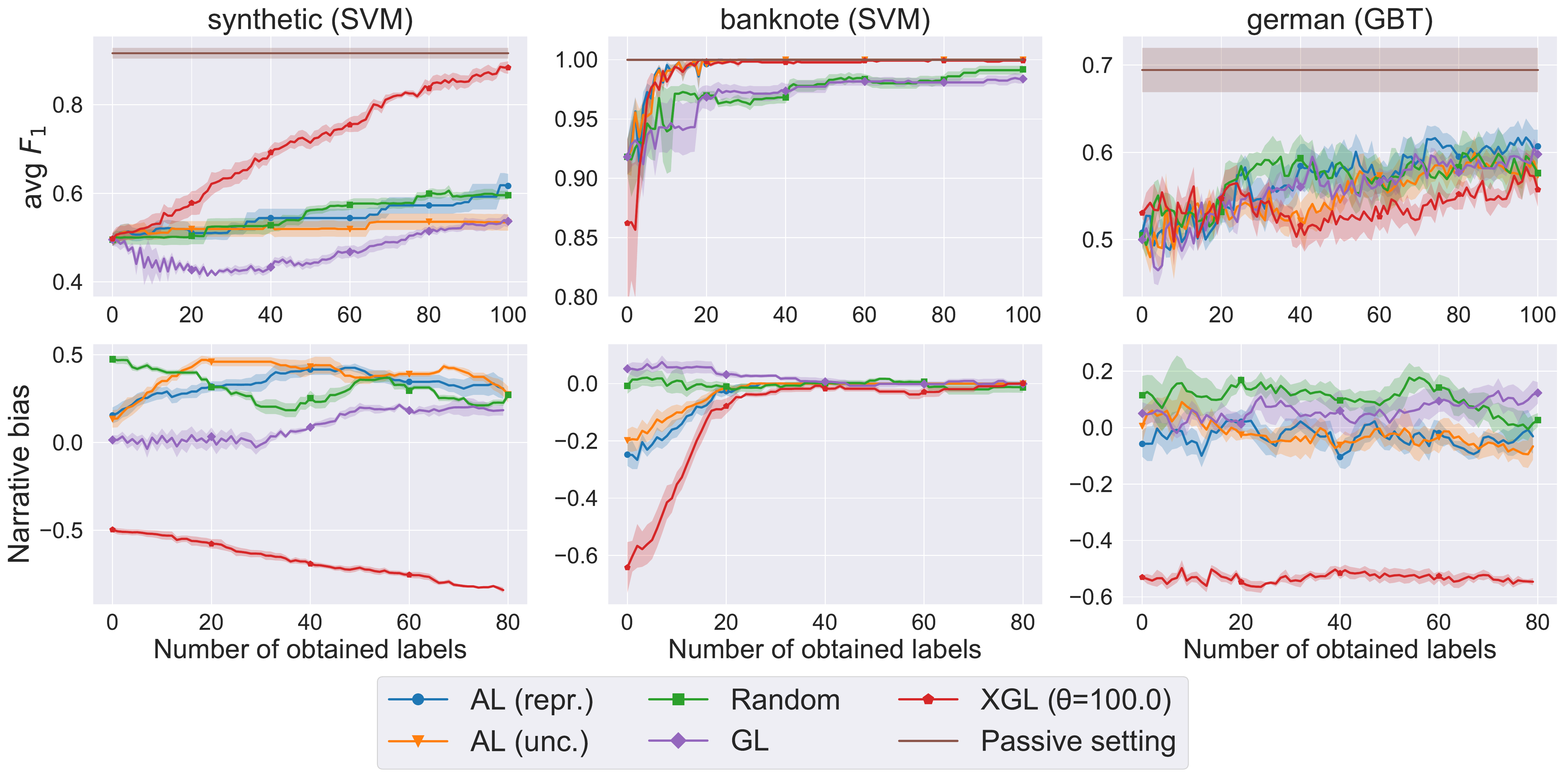}
    \caption{$F_1$ score (top) and narrative bias (bottom, the lower the better) of all competitors for increasing number of queries on three representative data sets:  the synthetic task (left), banknote (middle), and german (right).}
    \label{fig:f1_and_nb}
\end{figure*}

\textbf{Implementation:}  \method constructs global explanations by extracting rules from a decision tree classifier.\footnote{This simple approach outperformed more advanced ensemble-based rule learning algorithms in our experiments.}  The number of rules depends on the specific data set, but it usually ranges between 5 and 30. The accuracy of the tree w.r.t. the underlying classifier is high initially and drops slightly as the complexity increases, but usually remains above 80\% $F_1$. The simulator prioritizes mistakes occurring in the support of rules with the lowest $F_1$ score w.r.t. the ground-truth. A random mistake satisfying the chosen rule is returned for labelling. Naturally, human annotators can be imprecise when identifying rules with high loss.  To account for this, the simulator picks a rule with probability $P(\text{rule $i$}) = \exp(\theta m_i) / \sum_j \exp(\theta m_j)$ where $m_i$ is one minus the $F_1$ score of the $i$th rule and $\theta > 0$ models the supervisor's attention:  the larger $\theta$, the more likely a low-$F_1$ rule is chosen.  In most experiments we fix $\theta = 100$, which simulates an attentive and helpful annotator;  the effect of less attentive users with $\theta \in \{1, 10\}$ is studied in the final experiment.  If the chosen rule covers no mistake the process repeats, and if no rule makes any mistakes a random example is returned.

\textbf{Competitors:}  \method was compared against:
1)~GL:  guided learning, a human-initiated learning strategy in which the query instances are chosen randomly by balancing between the positive and negative classes, as per~\cite{attenberg2010label};
2)~AL unc:  uncertainty-based active learning, which picks an instance with the least difference between the probability of the two classes;
3)~AL repr:  active learning that favors instances that are both uncertain \emph{and} representative of the rest of the data;  the unlabelled instance with the lowest combination of uncertainty and average similarity to the other instances is chosen, see Eq.~13 in~\cite{fu2013survey};  changing the trade-off parameter $\beta$ had little effect, so it was fixed to $1$;
3)~Random sampling, a simple baseline that is hard to beat in practice;
4)~Passive learning, i.e., training on the whole training set.

\textbf{Experimental setup:}   Performance is computed on the test set using macro-averaged $F_1$ to give equal importance to each class.  The NB of the competitors was measured by taking the difference between the $F_1$ on the queries and the test set at the same iteration.  For random sampling, AL, XAL, and GL, a large NB value indicates that the queries over-estimate the model's performance, while a negative value indicates under-estimate.  Notice however that NB focuses on individual instances, whereas XGL presents the user a complete global explanation.  Hence for XGL a low value of NB simply means that the global explanation is useful for identifying bad mistakes and therefore to not mis-represent the quality of the learned model.

\begin{table*}[tb]
  \centering
  \begin{footnotesize}
  \begin{tabular}{lllllllll}
    \toprule
      &
      \multicolumn{2}{c}{\textbf{AL (repr.)}} &
      \multicolumn{2}{c}{\textbf{AL (unc.)}} &
      \multicolumn{2}{c}{\textbf{GL}} &
      \multicolumn{2}{c}{\textbf{XGL}} \\
      \cmidrule(lr){2-3}
      \cmidrule(lr){4-5}
      \cmidrule(lr){6-7}
      \cmidrule(lr){8-9}
      \textbf{Dataset} &
                {$F_1$} & {NB} & 
                {$F_1$} & {NB} &
                {$F_1$} & {NB} & 
                {$F_1$} & {NB} \\
        \midrule
        synthetic           & $0.55 \pm 0.03$ & 0.30 & $0.52 \pm 0.01$ & 0.34 & $0.47 \pm 0.04$ & 0.06 & $0.70 \pm 0.12\,\bullet$ & -0.69$\,\bullet$ \\ 
        \midrule
        adult               & $0.66 \pm 0.04$ & -0.17 & $0.67 \pm 0.02\,\bullet$ & -0.15 & $0.66 \pm 0.05$ & 0.08 & $0.64 \pm 0.06$ & -0.64$\,\bullet$ \\
        australian          & $0.80 \pm 0.06$ & -0.28 & $0.81 \pm 0.06$ & -0.31 & $0.79 \pm 0.06$ & 0.01 & $0.83 \pm 0.07\,\bullet$ & -0.83$\,\bullet$ \\
        banknote            & $0.99 \pm 0.04\,\bullet$ & -0.07 & $0.99 \pm 0.04\,\bullet$ & -0.08 & $0.97 \pm 0.04$ & 0.00 & $0.99 \pm 0.04\,\bullet$ & -0.19$\,\bullet$ \\ 
        cancer              & $0.95 \pm 0.03$ & -0.19 & $0.96 \pm 0.03\,\bullet$ & -0.18 & $0.93 \pm 0.03$ & 0.01 & $0.95 \pm 0.02$ & -0.46$\,\bullet$ \\
        credit              & $0.61 \pm 0.02$ & -0.08 & $0.61 \pm 0.02$ & -0.10 & $0.58 \pm 0.02$ & 0.06 & $0.64 \pm 0.01\,\bullet$ & -0.64$\,\bullet$ \\ 
        german              & $0.59 \pm 0.03\,\bullet$ & -0.07 & $0.55 \pm 0.03$ & -0.04 & $0.59 \pm 0.02\,\bullet$ & 0.02 & $0.53 \pm 0.02$ & -0.53$\,\bullet$ \\ 
        glass               & $0.77 \pm 0.03$ & -0.11 & $0.79 \pm 0.03\,\bullet$ & -0.06 & $0.77 \pm 0.03$ & 0.02 & $0.77 \pm 0.03$ & -0.47$\,\bullet$ \\ 
        heart               & $0.69 \pm 0.08$ & -0.23 & $0.70 \pm 0.06$ & -0.18 & $0.69 \pm 0.06$ & 0.01 & $0.71 \pm 0.07\,\bullet$ & -0.71$\,\bullet$ \\ 
        hepatitis           & $0.64 \pm 0.05$ & 0.09 & $0.66 \pm 0.07$ & 0.08 & $0.67 \pm 0.06$ & 0.05 & $0.68 \pm 0.05\,\bullet$ & -0.22$\,\bullet$ \\ 
        iris                & $0.94 \pm 0.02$ & -0.03 & $0.95 \pm 0.01\,\bullet$ & -0.01 & $0.94 \pm 0.01$ & -0.00 & $0.94 \pm 0.02$ & -0.08$\,\bullet$ \\ 
        magic               & $0.66 \pm 0.05\,\bullet$ & -0.15 & $0.65 \pm 0.06$ & -0.17 & $0.66 \pm 0.03\,\bullet$ & 0.04 & $0.64 \pm 0.04$ & -0.64$\,\bullet$ \\ 
        phoneme             & $0.69 \pm 0.07$ & -0.16 & $0.71 \pm 0.04\,\bullet$ & -0.17 & $0.68 \pm 0.05$ & 0.03 & $0.63 \pm 0.04$ & -0.63$\,\bullet$ \\ 
        plate-faults        & $0.65 \pm 0.06$ & -0.07 & $0.62 \pm 0.08$ & 0.01 & $0.66 \pm 0.08\,\bullet$ & 0.09 & $0.66 \pm 0.07\,\bullet$ & -0.65$\,\bullet$ \\ 
        risk                & $0.90 \pm 0.05$ & 0.08 & $0.95 \pm 0.08$ & -0.20 & $0.96 \pm 0.07\,\bullet$ & -0.00 & $0.96 \pm 0.07\,\bullet$ & -0.37$\,\bullet$ \\
        wine                & $0.89 \pm 0.02$ & 0.03 & $0.90 \pm 0.02$ & 0.02 & $0.91 \pm 0.03$ & 0.03 & $0.95 \pm 0.02\,\bullet$ & -0.17$\,\bullet$ \\ 
            \midrule
        adult+uu               & $0.59 \pm 0.05$ & 1.92 & $0.60 \pm 0.04\,\bullet$ & 9.92 & $0.61 \pm 0.03$ & 4.04 & $0.58 \pm 0.03$ & -0.58$\,\bullet$ \\ 
        australian+uu          & $0.61 \pm 0.05$ & 3.11 & $0.68 \pm 0.03$ & 0.32 & $0.69 \pm 0.03$ & 1.44 & $0.72 \pm 0.08\,\bullet$ & -0.72$\,\bullet$ \\ 
        banknote+uu            & $0.85 \pm 0.04\,\bullet$ & 3.07 & $0.83 \pm 0.03$ & 2.05 & $0.63 \pm 0.09$ & 1.66 & $0.77 \pm 0.03$ & -0.77$\,\bullet$ \\ 
        cancer+uu              & $0.87 \pm 0.04\,\bullet$ & 0.32 & $0.86 \pm 0.04$ & 0.59 & $0.77 \pm 0.02$ & 17.89 & $0.86 \pm 0.04$ & -0.86$\,\bullet$ \\ 
        credit+uu              & $0.55 \pm 0.02$ & 3.90 & $0.58 \pm 0.04$ & 4.27 & $0.53 \pm 0.04$ & 13.09 & $0.61 \pm 0.05\,\bullet$ & -0.61$\,\bullet$ \\ 
        german+uu              & $0.50 \pm 0.02$ & 1.36 & $0.52 \pm 0.02\,\bullet$ & 5.32 & $0.52 \pm 0.01\,\bullet$ & 4.57 & $0.51 \pm 0.02$ & -0.51$\,\bullet$ \\ 
        glass+uu               & $0.73 \pm 0.06\,\bullet$ & 3.53 & $0.72 \pm 0.04$ & 6.04 & $0.68 \pm 0.04$ & 3.65 & $0.72 \pm 0.03$ & -0.62$\,\bullet$ \\ 
        heart+uu               & $0.67 \pm 0.04\,\bullet$ & 0.46 & $0.66 \pm 0.04$ & 2.01 & $0.61 \pm 0.03$ & 3.23 & $0.66 \pm 0.04$ & -0.66$\,\bullet$ \\ 
        hepatitis+uu           & $0.61 \pm 0.07$ & 3.46 & $0.62 \pm 0.05$ & 0.63 & $0.60 \pm 0.04$ & 1.12 & $0.65 \pm 0.03\,\bullet$ & -0.41$\,\bullet$ \\ 
        iris+uu                & $0.75 \pm 0.02$ & 6.20 & $0.73 \pm 0.02$ & 4.82 & $0.75 \pm 0.02$ & 4.45 & $0.76 \pm 0.03\,\bullet$ & 4.04$\,\bullet$ \\ 
        magic+uu               & $0.59 \pm 0.03\,\bullet$ & 3.85 & $0.59 \pm 0.04\,\bullet$ & 3.27 & $0.59 \pm 0.03\,\bullet$ & 2.35 & $0.58 \pm 0.02$ & -0.58$\,\bullet$ \\ 
        phoneme+uu             & $0.61 \pm 0.06$ & 5.94 & $0.63 \pm 0.05$ & 4.70 & $0.65 \pm 0.04\,\bullet$ & 11.88 & $0.65 \pm 0.03\,\bullet$ & -0.65$\,\bullet$ \\ 
        plate-faults+uu        & $0.60 \pm 0.04$ & 2.94 & $0.58 \pm 0.07$ & 5.64 & $0.61 \pm 0.03\,\bullet$ & 3.41 & $0.61 \pm 0.04\,\bullet$ & -0.61$\,\bullet$ \\ 
        risk+uu                & $0.87 \pm 0.05$ & 5.68 & $0.92 \pm 0.05\,\bullet$ & 8.80 & $0.91 \pm 0.03$ & 11.48 & $0.88 \pm 0.06$ & -0.88$\,\bullet$ \\ 
        wine+uu                & $0.80 \pm 0.04$ & 10.26 & $0.80 \pm 0.03$ & 9.60 & $0.74 \pm 0.05$ & 9.67 & $0.81 \pm 0.04\,\bullet$ & 2.15$\,\bullet$ \\
        \bottomrule
  \end{tabular}
  \end{footnotesize}
  \caption{Results for all methods on the original and UU-augmented data sets.  The $F_1$ and NB are averaged across all iterations and folds.  The best average values for each data set are indicated by a bullet.
  }
  \label{tab:results-table}
\end{table*}

\textbf{Narrative Bias:}  Table~\ref{tab:results-table} reports the results for
all methods and data sets.  The results show that \method attains consistently
lower NB than the competitors.  This is particularly clear in the synthetic
data set and for the ``+uu'' data set variants in which we injected high-cost
disjoint clusters.  In these cases, all methods except ours over-estimate the
performance of the classifier while XGL has negative NB (until the classifier
approaches zero loss).  This means that the rule sets extracted by \method are
consistently effective in identifying regions of low performance.
The $F_1$ and NB curves for three representative datasets are reported in
Figure~\ref{fig:f1_and_nb}.  The query $F_1$ was slightly smoothed for
readability.
In synthetic, all methods suffer from NB at all iterations except for XGL and
GL for a few iterations.  The competitors underperform because: i)~AL focuses
on the known mistakes while ignoring the UUs, as illustrated in
Figure~\ref{fig:example}, ii)~GL attempts to acquire a balanced data set and
over-samples the minority class, as illustrated in the Supplementary Material.
The random baseline behaves similarly.
The competitors however perform well on the other data sets, because the
ground-truth does not include disjunctive concepts, as shown by banknote in the
Figure.
The situation changes substantially for the ``+uu'' data sets in which we
injected high-cost disjoint clusters.  In this challenging case, all approaches
except \method have NB $\gg 0$.  This shows that, so long as the classifier
suffers from UUs, XGL shields it from narrative bias, and allows us to answer
\textbf{Q1} in the affirmative.

\textbf{Predictive Performance:}  \method produces predictions on par or better
than those of the competitors in most datasets.  On synthetic, which is
particularly hard, the performance difference is quite marked, with XGL
outperforming the competitors by almost $20\%$ $F_1$.  This is again due to
UUs:
%
%
AL and random sampling only rarely query instances from the red class, which is
the reason for their slow progress shown in Figure~\ref{fig:f1_and_nb} (left),
while GL over-samples the minority class.
XGL performs similarly or outperforms all competitors in all original datasets
and in all ``+uu'' variants.  The most problematic case is german, where XGL
tends to perform poorly in terms of F1 regardless of choice of base classifier (SVM, gradient
boosting;  results not shown), however still performs best in terms of narrative bias.
Summarizing, the results show that in the presence of UUs XGL tends to learn
better classifiers compared to the alternatives, while if UUs are less of an
issue, XGL performs reasonably anyway.

\begin{figure*}[tb]
    \centering
        \includegraphics[width=0.8\linewidth]{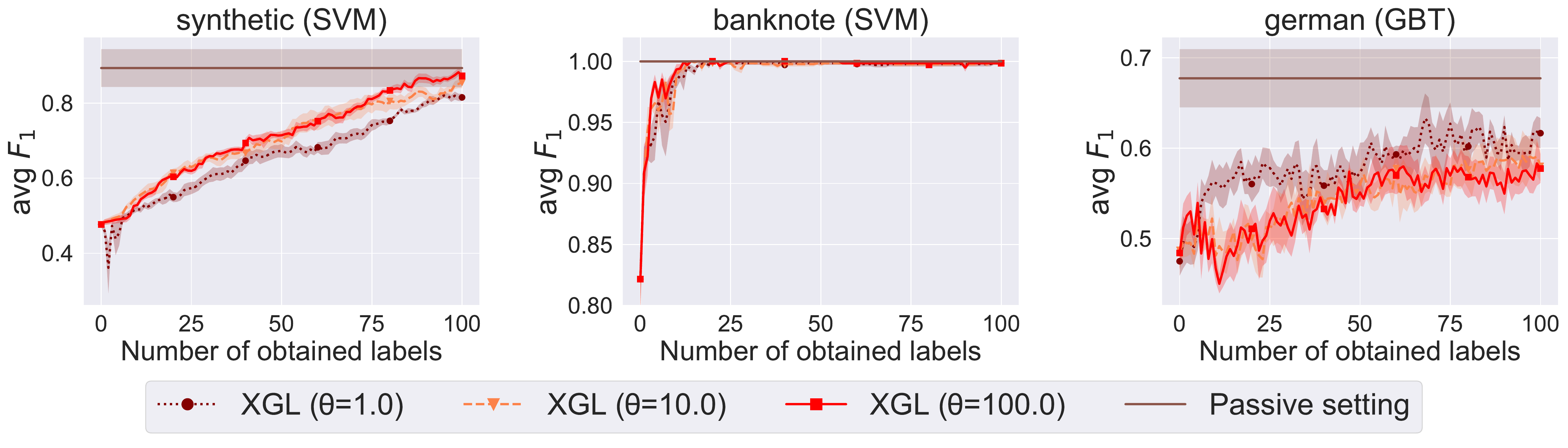}
    \caption{Left:  Predictive performance of XGL with varying $\theta$.}
    \label{fig:methods_and_thetas}
\end{figure*}

\textbf{Answering Q3:}  The results obtained by varying the attention parameter
$\theta$ are presented in Figure~\ref{fig:methods_and_thetas} (right).  For
$\theta = 100$, as used in the previous experiments, the simulator selects a
rule with minimal $F_1$ out of the ones in the global explanation about $90\%$
of the time.  Reducing $\theta$ introduces more randomness in the choice:  for
$\theta = 10$ the simulator chooses the worst rule only $50$-$80\%$ of the
time.  Hence, the provided examples are not as effective in correcting the
classifier's mistakes.  Even with this less attentive user, however, XGL
manages to achieve predictive performance close to that of the most attentive
user ($\theta = 100$).  Lowering $\theta$ by one order of magnitude, which
corresponds to selecting a non-pessimal rule more than $50\%$ of the time, the
performance of XGL does not decrease substantially in synthetic and banknote,
while for german performance does increase.  This is consistent with the fact
that random sampling does perform better than \method in these two data sets
(see the Supplementary Material.)  These results show that XGL works even for
inattentive, sub-optimal supervisors, as long as they manage to select
informative mistakes often enough.

\section{Related Work}

Our work is motivated by explanatory interactive learning~\cite{teso2019explanatory,schramowski2020right}, of which both explanatory guided learning and explanatory active learning are instances.

Human-initiated interaction has been proposed as a strategy for handling unknown unknowns~\cite{bansal2019beyond,vandenhof2019contradict,attenberg2015beat}.  XGL takes inspiration from guided learning (GL)~\cite{attenberg2010label,simard2014ice,beygelzimer2016search}.  In GL, the human supervisor is asked to supply examples of the minority class, but the protocol is black-box:  no guidance (besides a textual description of the class) is given, hence the user has no clue of what the predictor believes and may have trouble providing non-redundant examples.  Our experiments support this observation.  Still, GL proved useful in combating label skew in real-world classification tasks in industrial deployments.  Given how close the XGL and GL interaction protocols are, this arguably shows that XGL can also be implemented at industrial scale.  Other human-initiated approaches also rely on opaque interaction~\cite{attenberg2015beat,vandenhof2019contradict}.  Our insight that human-initiated interaction has to and should be combined with global explanations to combat narrative bias is novel.

XGL complements recent work in interactive machine learning showing that only interactive teachers can successfully teach to black-box classifiers~\cite{melo2018interactive,chen2018near,dasgupta2019teaching}.  These results require the teacher to have access to the whole decision surface of the learner, which human supervisors cannot do.  Our work identifies global explanations as a suitable solution.

Most closely related to our work, Lakkaraju et al. combine multi-armed bandits (MABs) and user interaction to identify interpretable clusters with high UU content~\cite{lakkaraju2017identifying}.  While their combination of interaction and interpretable clusters offers support for XGL, the two have different goals: XGL aims to learn from UUs and other examples while their approach focuses on identifying the UUs (only) of a given classifier without any learning.  One idea is to wrap this approach into XGL so to aid the supervisor in the search for UUs.  However in XGL the model changes at all iterations, rendering MABs unsuitable.  While the idea of aiding the user in this sense is sensible (as discussed above), extending MABs to our drifting case, although possible~\cite{liu2020incentivized}, is non-trivial and left to future work.

The concept of guidance has been used in the literature to mean different things.  In XGL, the machine guides the user with global explanations.  This is related to teaching guidance~\cite{cakmak2014eliciting} in which the machine educates the supervisor by providing instructions as to how to select informative examples in non-technical terms.  For instance, the machine could guide the user in choosing instances close to the decision boundary.  This kind of guidance could be fruitfully combined with XGL to identify more informative mistakes within individual rules.  This non-trivial extension is left to future work.

\section{Conclusion}

This work identifies the issue of narrative bias in current explanatory interactive learning and addresses it by combining machine guidance, in the form of global explanations, and human-initiated interactive learning.  Global explanations were validated theoretically by leveraging a novel link to interactive machine teaching.  Our empirical analysis has showcased the advantages of our approach over alternative machine- and human-initiated interactive learning strategies.

We view XGL as a stepping stone for further research.  In particular, XGL can and should be improved by:
i)~integrating search support technology to assist users in exploring the global explanation,
ii)~intelligently combining machine- and human-initiated interaction to lower the cognitive cost of global explanations, and
iii)~introducing adaptive multi-resolution explanations so to enable the user to ``zoom in'' ambiguous regions and provide better supervision,
Many other improvements are possible.  These developments are left to future work.

\bibliography{paper}

\newpage
\section{Appendix A:  Proof of Proposition 1}

\begin{algorithm}[tb]

    \begin{algorithmic}[1]
        \State $S \gets \emptyset$, \; $w(\vx) \gets |\calX|^{-1}$, \; $\tau(\vx) \sim \exp(\lambda_{\calG}) $
        \For{$t = 1, 2, \ldots$}
            \State machine supplies $g$ learned on $S$
            \State determine $\Delta(g) := \{ \vx \in \calX : g(\vx) \ne g^*(\vx) \}$
            \If{$|\Delta(g)| \le \eta$}
                \State \textbf{break}
            \EndIf
            \While{$\sum_{\vx \in \Delta(g)} w(\vx) < 1$}
                \For{$\vx \in \Delta(g)$}
                    \State $w(\vx) \gets 2 \cdot w(\vx)$
                    \If{$w(\vx) > \tau(\vx) \;\, \land \not\exists \, y \, . \, (\vx, y) \in S$}
                        \State $S \gets S \cup \{ (\vx, g^*(\vx)) \}$
                    \EndIf
                \EndFor
            \EndWhile
            \State present $S$ to the machine
        \EndFor
    \end{algorithmic}

    \caption{\label{alg:teacher} Teaching oracle: $g^* \in \calG$ is the
    ground-truth, $g \in \calG$ is a candidate global explanation, $S$ is the training set, $\eta$ is the target tolerance.}

\end{algorithm}

The proof is split into Lemmas~\ref{lemma:1} to \ref{lemma:4}, which are lifted directly from~\cite{dasgupta2019teaching} and reported here for completeness.  The proofs have been rephrased for readability and to  and very slightly modified to work on the space of global explanations ($\calG$) instead of hypothesis space ($\calH$).  See Algorithm~\ref{alg:teacher} for the pseudo-code of the teaching oracle.

\begin{lemma}
\label{lemma:1}
The weight of any $\vx \in \calX$ doubles at most $\lg 2|\calX|$ times.
\begin{proof}
The weight $w(\vx)$ starts at $\frac{1}{|\calX|}$.  Since it only increases (doubles) when $\vx$ belongs to $\Delta(g)$, which is a subset of $\calX$ with total weight at most $1$, it grows at most to $2$.  Hence, $\frac{1}{|\calX|} \cdot 2^s < 2$, where $s$ is the number of steps.
\end{proof}
\end{lemma}

\begin{lemma}[Number of doublings]
\label{lemma:2}
The total number of doublings performed by the oracle in Algorithm~\ref{alg:teacher} is at most $|S(g^*) | \lg 2|\calX|$.  Moreover, it always holds that
$
    \textstyle
    \sum_{\vx \in \calX} w(\vx) \le 1 + |S(g^*)| \lg 2|\calX|
$.
\begin{proof}
A doubling only occurs if $g \ne g^*$.  In this case, $\Delta(g)$ must intersect $S(g^*)$, otherwise $S(g^*)$ would not disambiguate $g$ from $g^*$.  Therefore, doubling $\Delta(g)$ doubles at least one element of $S(g^*)$.  But by Lemma~\ref{lemma:1} the elements of $S(g^*)$ cannot be doubled more than $|S(g^*)| \lg 2|\calX|$ times overall.

Since the summation $\sum_{\vx \in \calX} w(\vx)$ starts at a value less than $1$, each doubling step increases it by at most $1$. 
The lemma follows by noting that initially the sum satisfies $\sum_{\vx \in \Delta(g)} w(\vx) < 1$ and there are at most $|S(g^*)| \lg 2|\calX|$ doubling steps.
\end{proof}
\end{lemma}

\begin{lemma}[Expected number of examples]
\label{lemma:3}
In expectation over the choice of $\tau$, once the algorithm terminates, the number of examples in $S$ is at most $(1 + |S(g^*)|\lg 2|\calX|) \ln(|\calG| / \delta)$.
\begin{proof}
Let $w(\vx)$ be the weight at termination.  The probability that a given $\vx$ has been added to $S$ is:
\begin{align*}
    \Pr(w(\vx) > \tau(\vx))
        & \; = 1 - \Pr(\tau(\vx) > w(\vx)) \\
        & \; = 1 - \exp(-\lambda w(\vx)) \\
        & \; \le \lambda w(\vx)
\end{align*}
where $\lambda = \ln(|\calG| / \delta)$.  Therefore,
\begin{align*}
    \bbE_{\tau}[|S|]
        & \textstyle = \sum_{\vx \in \calX} \Pr(w(\vx) > \tau(\vx)) \\
        & \textstyle \; \le \sum_{\vx \in \calX} \lambda w(\vx) \\
        & \textstyle \; \le \lambda(1 + |S(g^*)| \lg 2 |\calX|)
\end{align*}
The first step holds because $\tau$ is sampled i.i.d. for each $\vx$ and the last one because of Lemma~\ref{lemma:2}.
\end{proof}
\end{lemma}

\begin{lemma}
\label{lemma:4}
At the end of each iteration, all hypotheses $g' \in \calG$ with $g' \ne g^*$ and total weight $\sum_{\vx \in \Delta(g')} w(\vx) \ge 1$ are invalidated by $S$ (i.e., $S$ contains an instance $\vx$ in which $g'(\vx) \ne g^*(\vx)$) with probability at least $1 - \delta$.
\begin{proof}
Fix $g' \ne g^*$ and consider the first point in time at which $\sum_{\vx \in \Delta(g')} w(\vx) \ge 1$.  Then:
\begin{align*}
        & \textstyle \Pr(\text{$g'$ is not invalidated by $S$})
    \\
        & \quad \textstyle = \Pr(\text{no point in $\Delta(g')$ is added to $S$})
    \\
        & \quad \textstyle = \prod_{\vx \in \Delta(g')} \Pr(w(\vx) \le \tau(\vx))
    \\
        & \quad \textstyle = \prod_{\vx \in \Delta(g')} \exp(-\lambda w(\vx))
    \\
        & \quad \textstyle = \exp(-\lambda \sum_{\vx \in \Delta(g')} w(\vx))
    \\
        & \quad \textstyle  \le \exp(-\lambda) = \frac{\delta}{|\calG|}
\end{align*}
Hence, the probability that some hypothesis $g \ne g^*$ is \emph{not} invalidated by $S$ is at most:
\begin{align*}
    & \textstyle \Pr(\exists \, g \in \calG \, . \, g \ne g^* \land \text{$h$ is not invalidated by $S$})
    \\
        & \textstyle \quad \le \sum_{g \in \calG : g \ne g^*} \Pr(\text{$g$ is not invalidated by $S$})
    \\
        & \textstyle \quad \le \sum_{g \in \calG : g \ne g^*} \frac{\delta}{|\calG|}
    \\
        & \textstyle \quad = \delta
\end{align*}
The first inequality follows from the union bound and the lemma follows by taking the complement.
\end{proof}
\end{lemma}

Combining Lemmas~\ref{lemma:2},~\ref{lemma:3}, and~\ref{lemma:4} and using the inequality 
$L(h, h^*) \le L(g, g^*) + 2\rho$ gives us Proposition~1.

\section{Appendix B:  Additional Implementation Details}

\subsection{Data}

A summary of the data sets is given in Table~\ref{tab:datasets-table}. In australian, credit, risk and hepatitis the categorical features are represented with numerical labels, while in adult, german and heart we encode them as one-hot vectors. The numerical features for all data sets are normalized either by standardization to zero mean and unit variance or by scaling between zero and one (in synthetic and hepatitis). The missing values are imputed using mode (adult) or median (hepatitis). The number of features for the credit data set are reduced to three using recursive feature elimination.

\begin{table}[h]
    \centering
    \begin{minipage}{\linewidth}
        \centering
        \begin{footnotesize}
            \begin{tabular}{lcccccc}
            \toprule
            {\bf Dataset} & {\bf Examples} & {\bf \#Pos} & {\bf \#Neg} & {\bf \#Cat} & {\bf \#Num} \\
            \midrule
            synthetic  & 1041   & 100    & 941    & 0  & 2   \\
            \midrule
            adult        & 48842  & 11687  & 37155  & 8  & 6  \\ 
            australian   & 690    & 383    & 307    & 8  & 6  \\ 
            banknote     & 1372   & 610    & 762    & 0  & 4  \\ 
            cancer       & 569    & 212    & 357    & 0  & 30 \\ 
            credit       & 30000  & 6636   & 23364  & 10 & 13 \\ 
            german       & 1000   & 300    & 700    & 13 & 7  \\ 
            glass        & 214    & 146    & 68     & 0  & 9  \\
            heart        & 303    & 139    & 164    & 7  & 7  \\ 
            hepatitis    & 155    & 32     & 123    & 13 & 6  \\
            iris         & 150    & 50     & 100    & 0  & 4  \\
            magic        & 19020  & 6688   & 12332  & 0  & 10 \\    
            phoneme      & 5404   & 1586   & 3818   & 0  & 5  \\
            risk         & 2912   & 1673   & 1239   & 2  & 34 \\    
            plate-faults & 1941   & 285    & 1656   & 0  & 27 \\
            wine         & 178    & 48     & 130    & 0  & 13 \\
            \bottomrule
        \end{tabular}
        \end{footnotesize}
        \caption{Data set summary:  number of examples (positive and negative) and attributes (categorical and numerical).}
        \label{tab:datasets-table}
    \end{minipage}
\end{table}

\subsection{Models}

For every data set, we compare several learning algorithms implemented in the scikit-learn library and choose the one that has the best passive performance.
To capture the complexity of the synthetic data set we use an SVM with RBF kernel and parameters $\gamma=100$ and $C=100$. The predictor in banknote and breast cancer experiments is RBF SVM with $\gamma=0.01$ and $C=100$. 
For the remaining experiments, gradient boosting trees with default parameters are used.

\subsection{Competitors}

The uncertainty-based AL strategy follows the least-certain strategy discussed in~\cite{settles2012active}.
The density-weighted AL strategy is implemented using Eq.~13 in~\cite{fu2013survey}, which consists of two terms. The first term computes the informativeness of $x$ according to a variant of the uncertainty sampling framework that queries the example for which the learner has the least confidence in the most likely prediction~\cite{settles2012active}. The second term is implemented using cosine similarity.

\section{Appendix C:  Additional Results}

\subsection{Selected Queries}

\begin{figure*}[p]
    \centering
    \captionsetup[subfigure]{justification=centering}
    \subfloat[XGL, \nth{10} iteration]
        {
        \includegraphics[width=.3\linewidth]{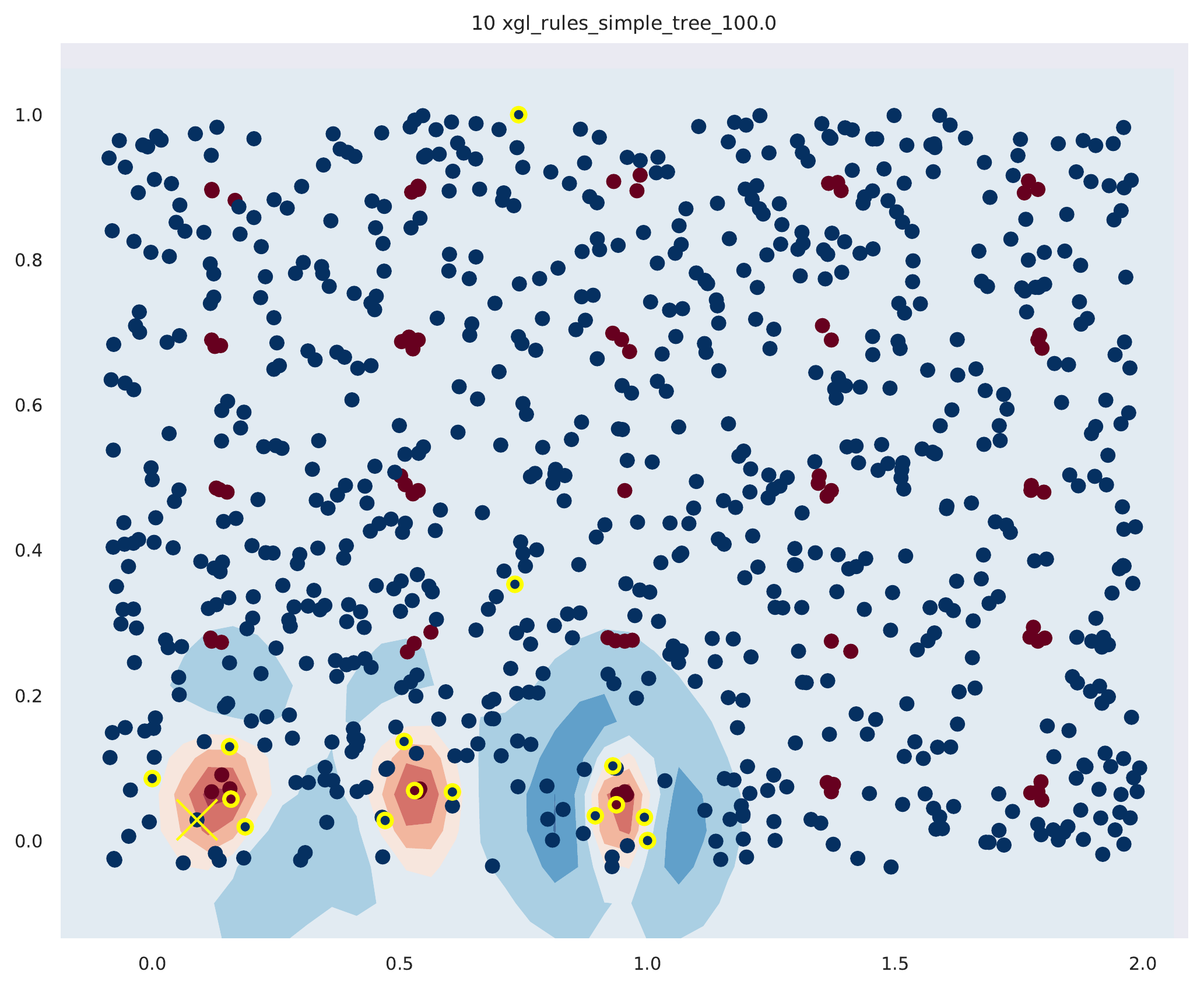}
        } \hfill
    \subfloat[XGL, \nth{50} iteration]
        {
        \includegraphics[width=.3\linewidth]{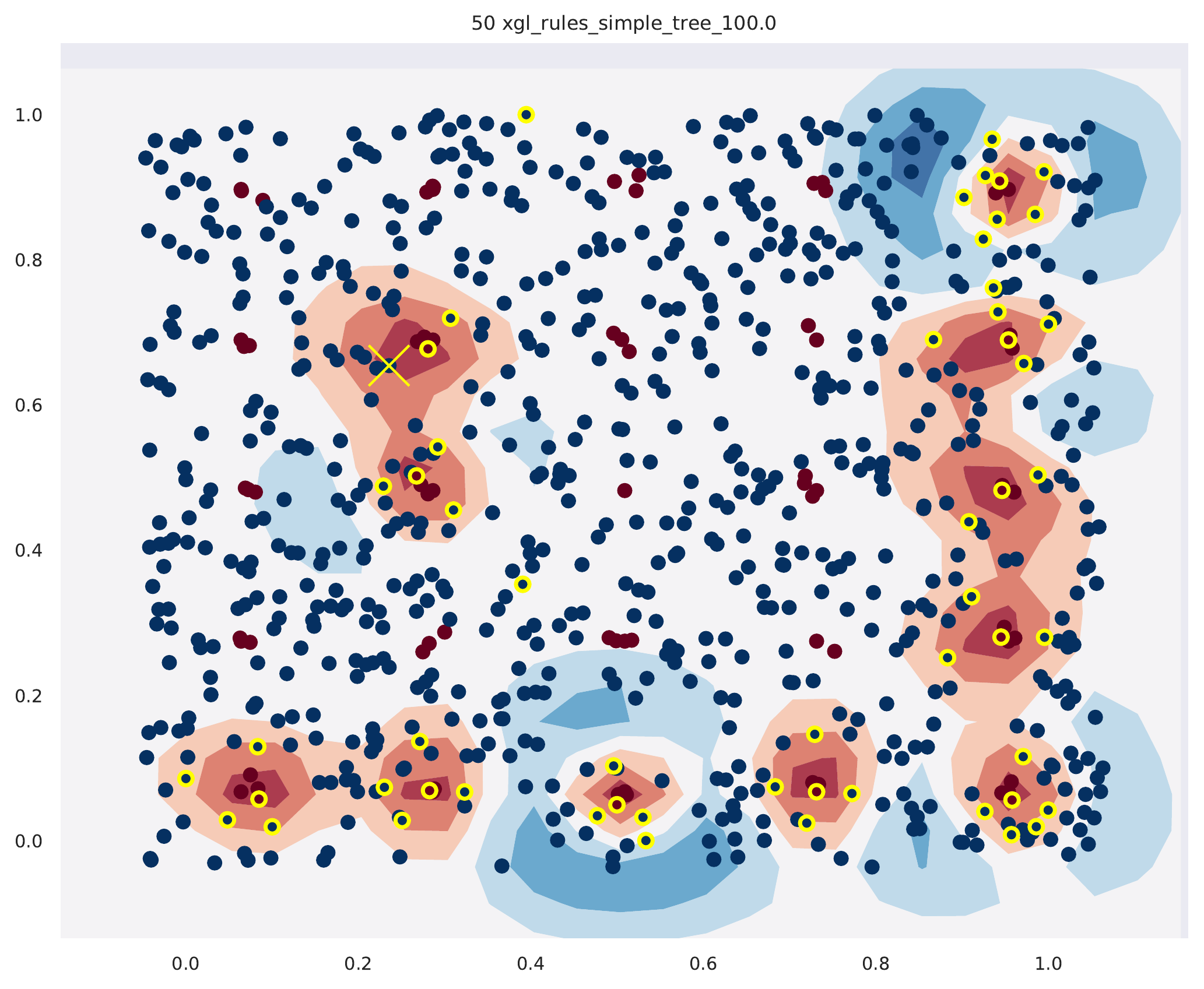}
        } \hfill
    \subfloat[XGL, \nth{100} iteration]
        {
        \includegraphics[width=.3\linewidth]{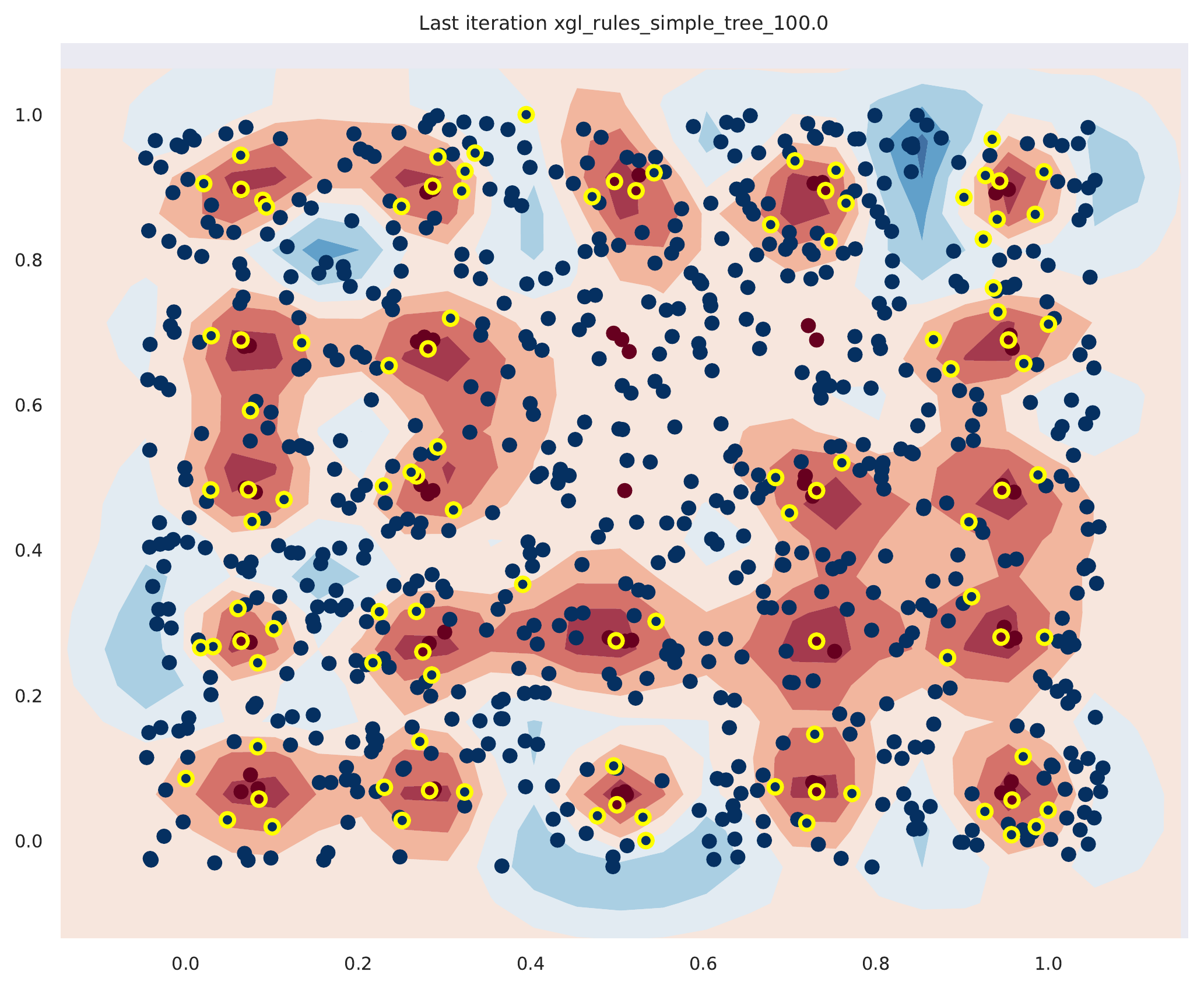}
        } \hfill
    \\
    \subfloat[AL (unc.), \nth{10} iteration]
        {
        \includegraphics[width=.3\linewidth]{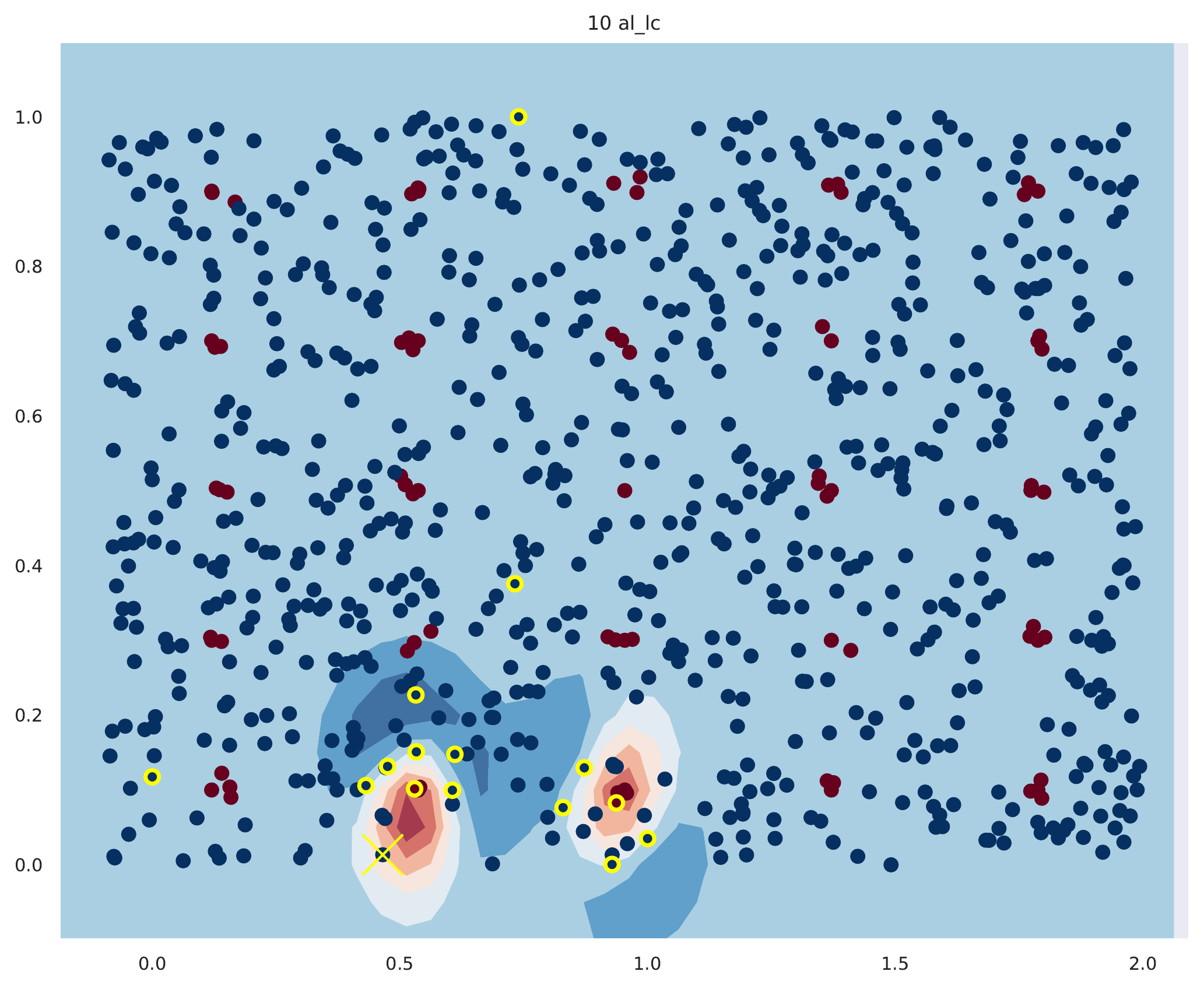}
        } \hfill
    \subfloat[AL (unc.), \nth{50} iteration]
        {
        \includegraphics[width=.3\linewidth]{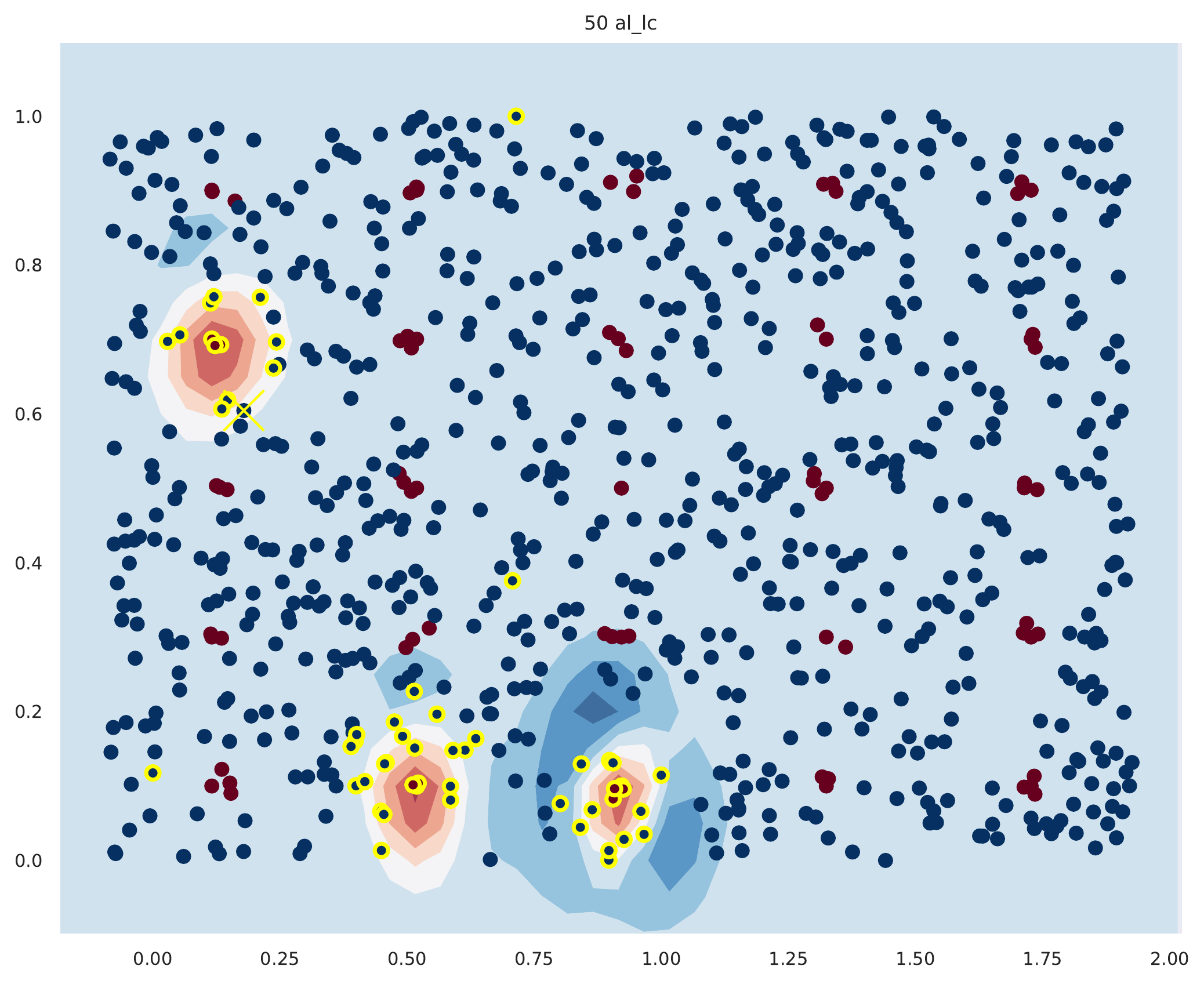}
        } \hfill
    \subfloat[AL (unc.), \nth{100} iteration]
        {
        \includegraphics[width=.3\linewidth]{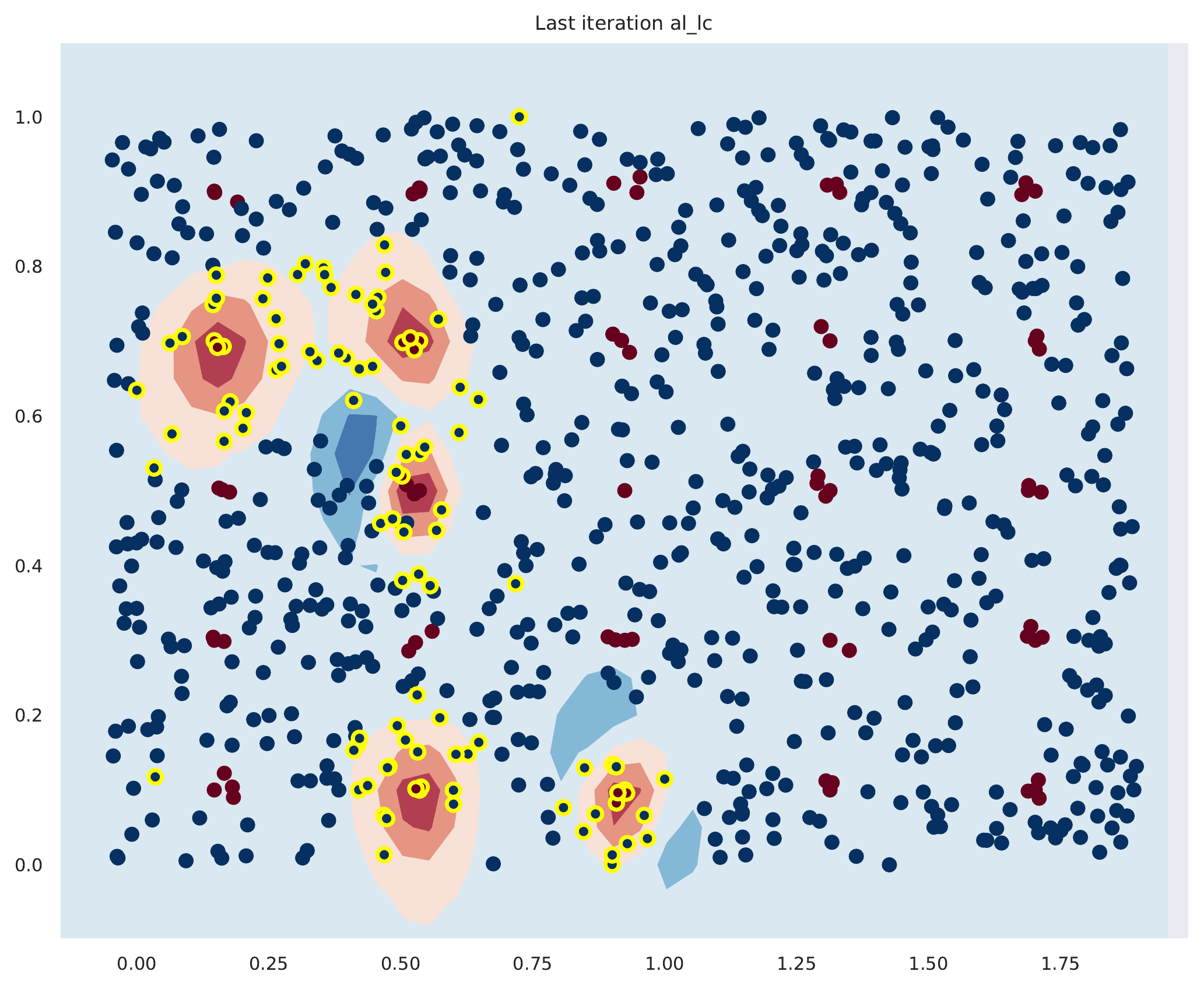}
        } \hfill
    \\
    \subfloat[AL (repr.), \nth{10} iteration]
        {
        \includegraphics[width=.3\linewidth]{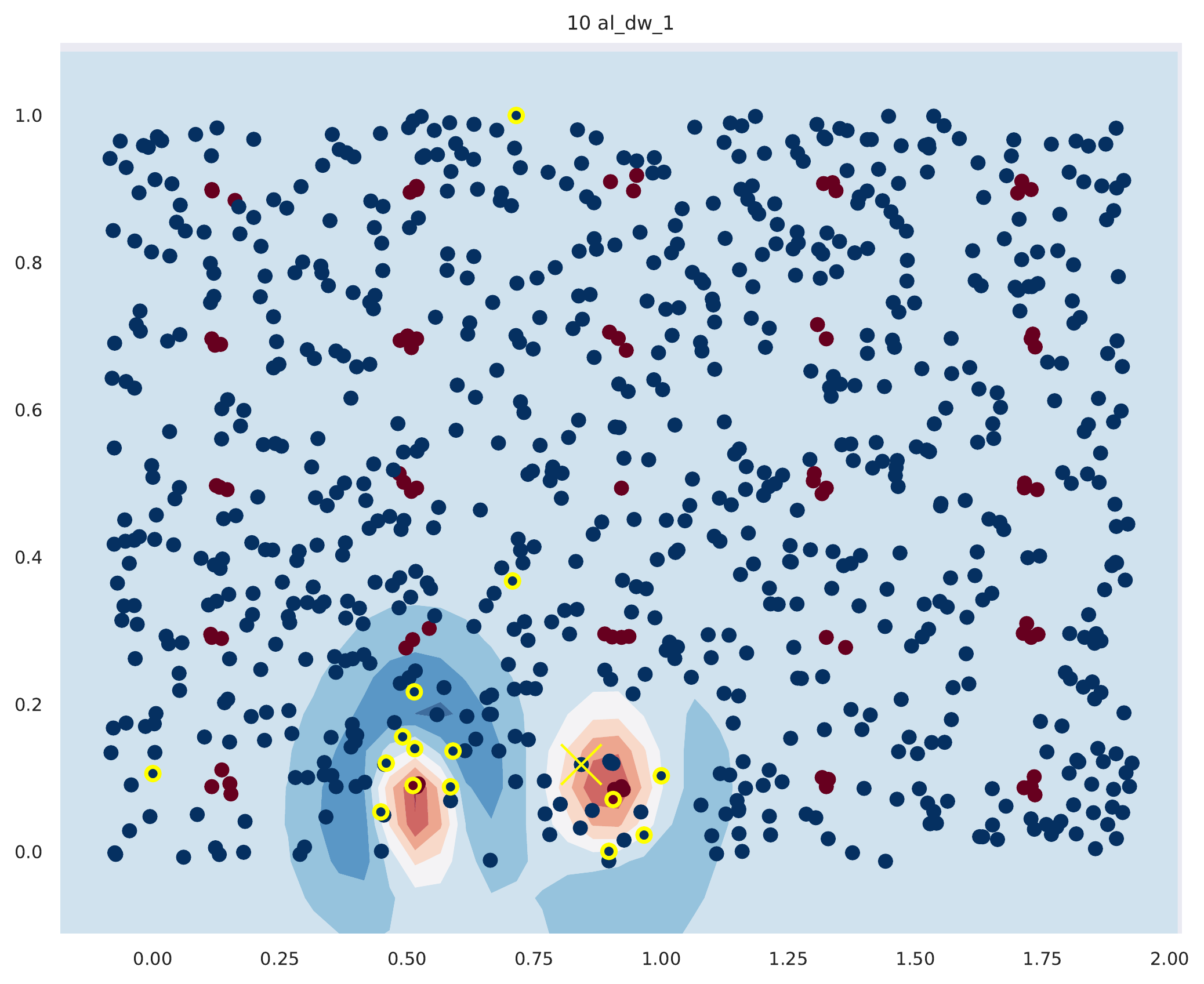}
        } \hfill
    \subfloat[AL (repr.), \nth{50} iteration]
        {
        \includegraphics[width=.3\linewidth]{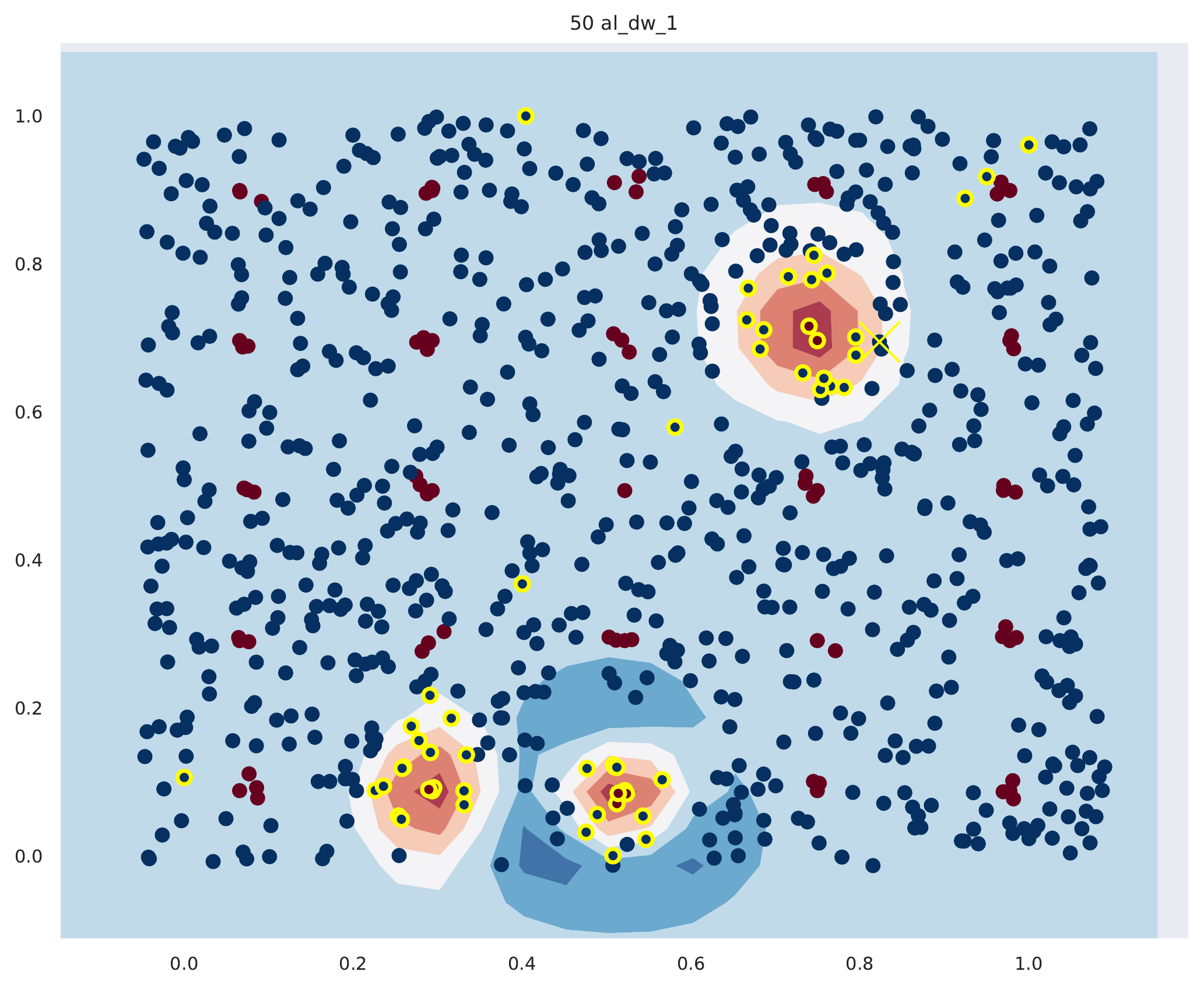}
        } \hfill
    \subfloat[AL (repr.), \nth{100} iteration]
        {
        \includegraphics[width=.3\linewidth]{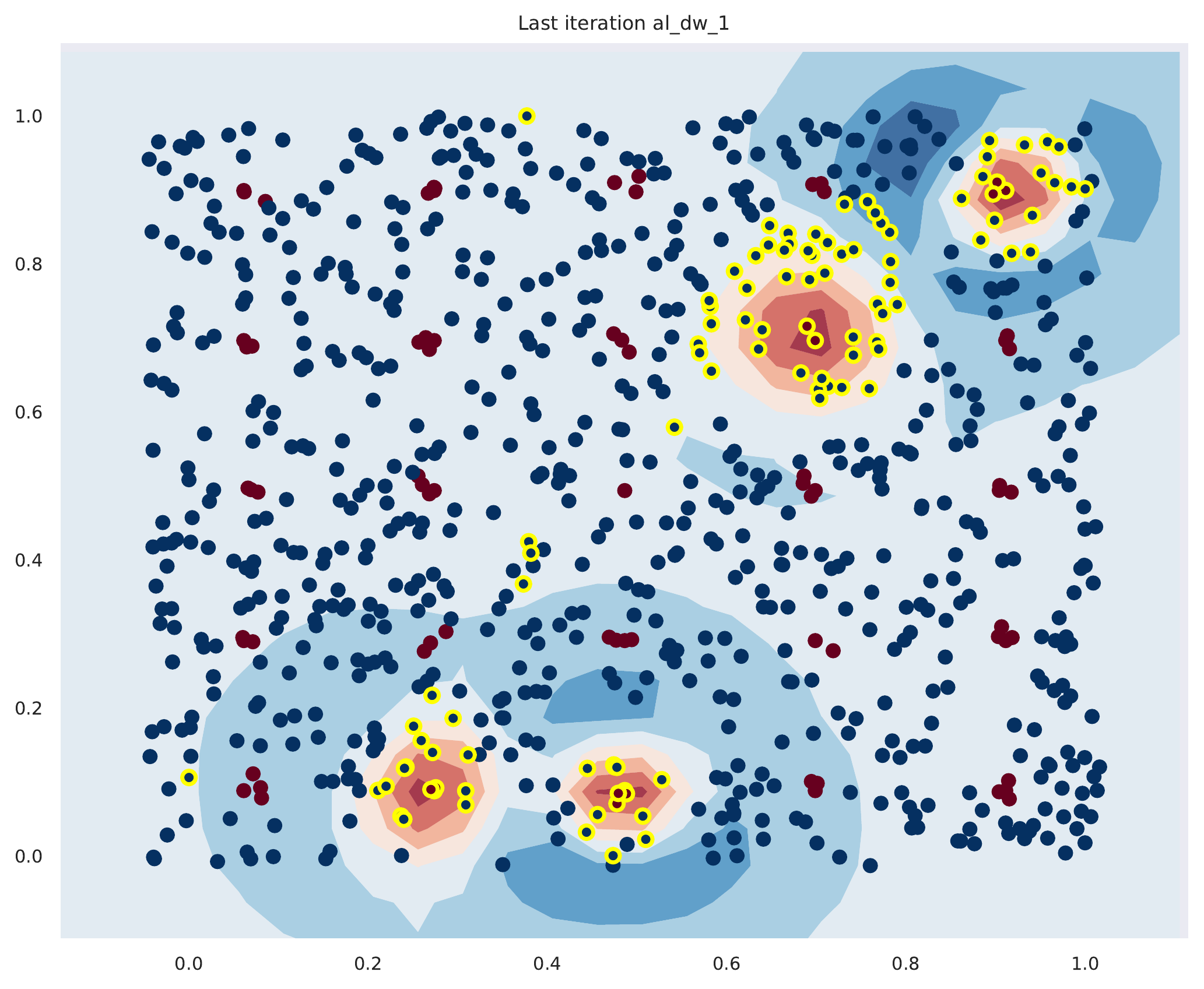}
        } \hfill
    \\
    \subfloat[GL, \nth{10} iteration]
        {
        \includegraphics[width=.3\linewidth]{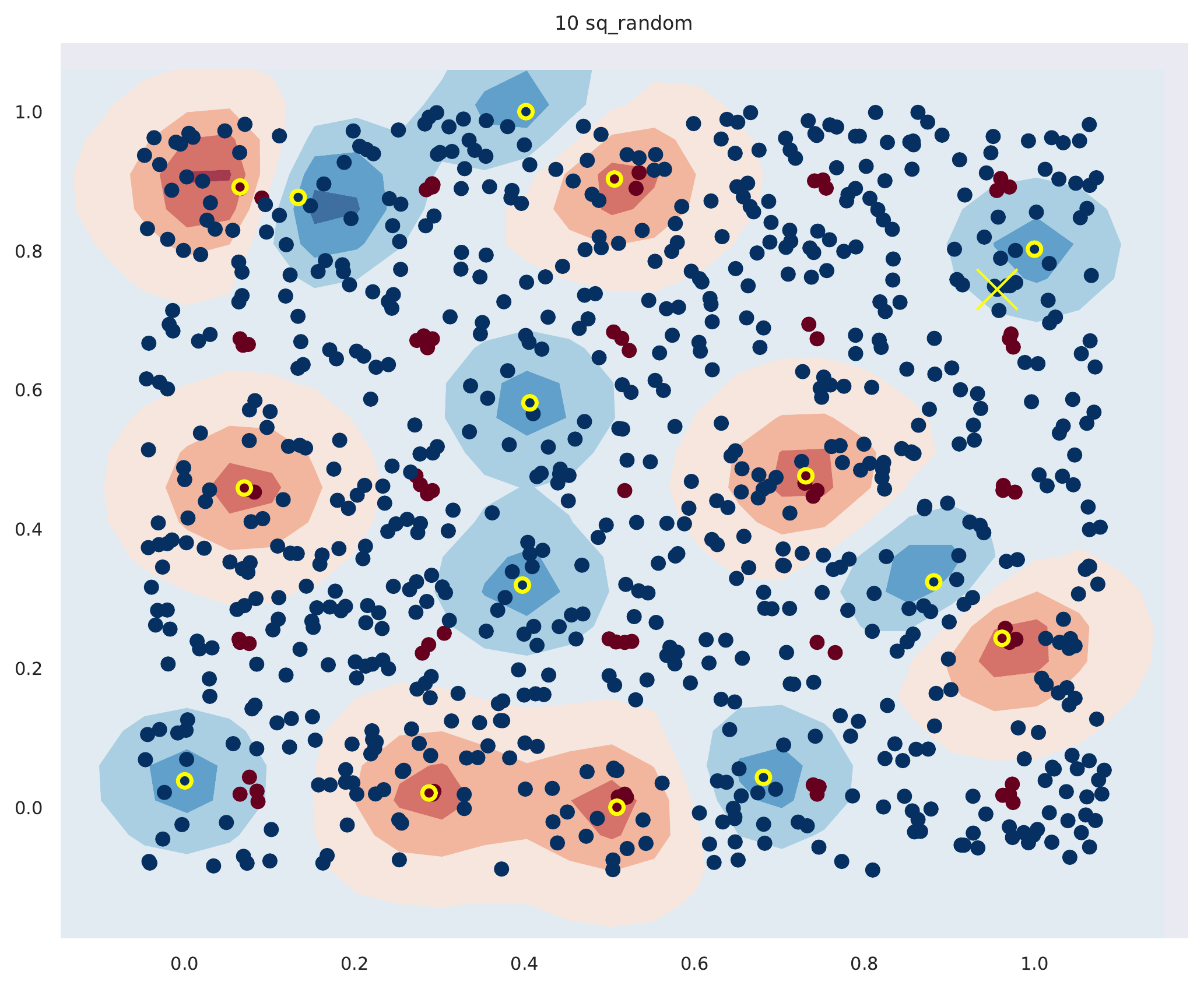}
        } \hfill
    \subfloat[GL, \nth{50} iteration]
        {
        \includegraphics[width=.3\linewidth]{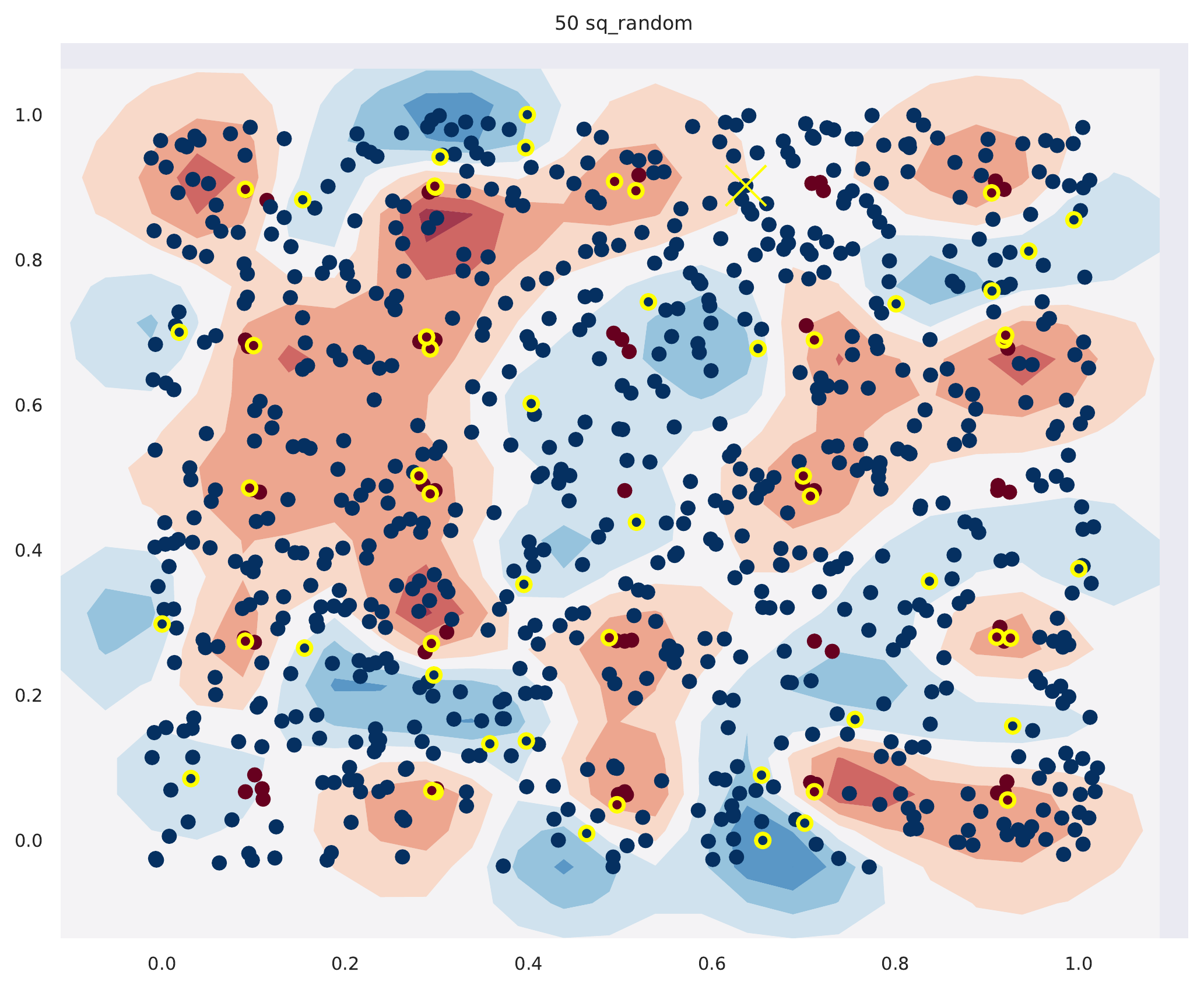}
        } \hfill
    \subfloat[GL, \nth{100} iteration]
        {
        \includegraphics[width=.3\linewidth]{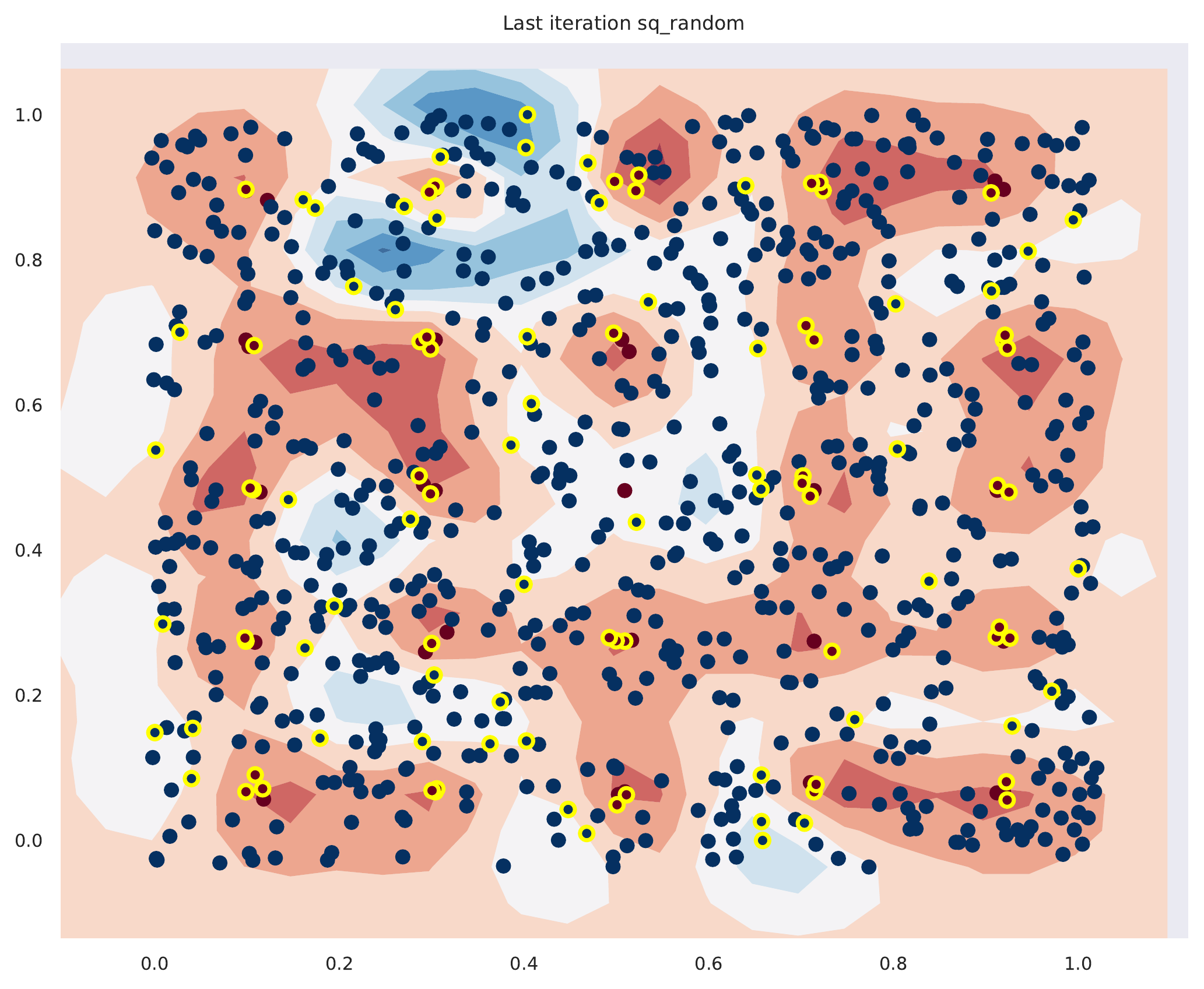}
        } \hfill
    \caption{Queries selected by XGL and AL on the synthetic task.  From top to bottom: XGL, uncertainty-based AL and density-weighted AL.}
    \label{fig:selected_queries}
\end{figure*}

The queries selected by XGL and by all competitors competitors on the synthetic data set are illustrated in Figure~\ref{fig:selected_queries}.

The exploitative nature of uncertainty sampling leads uncertainty-based AL to select instances around known red clusters, thus wasting querying budget on largely redundant instances, and ignoring the unknown red clusters (second row).
Combining uncertainty and representativeness does not improve the situation (third row), as density is not highly indicative of unknown red clusters.
This means that the explanatory narrative output by XAL using these two query selection strategies is not representative of the generalization ability of the predictor as it ignores the unknown unknowns.
In stark contrast, XGL enables a knowledgeable and helpful simulator to identify many of the unknown red clusters very quickly.  Indeed, XGL discovers most of the clusters in 140 iterations, while AL is stuck to just a few.

As for GL, recall that no explanations are shown to the supervisor.  It is evident that an uninformed supervisor is not unlikely to present the learner instances from regions where it is already performing well:  red points are selected from already found red clusters.  In XGL, the chosen instances are balanced between refining the decision boundary and exploring new red clusters.
This emphasizes the importance of global explanations from a sampling complexity perspective and validates XGL as an explainable generalization of GL.

\subsection{Narrative Bias}

\begin{figure*}[p]
    \centering
    \includegraphics[width=\linewidth]{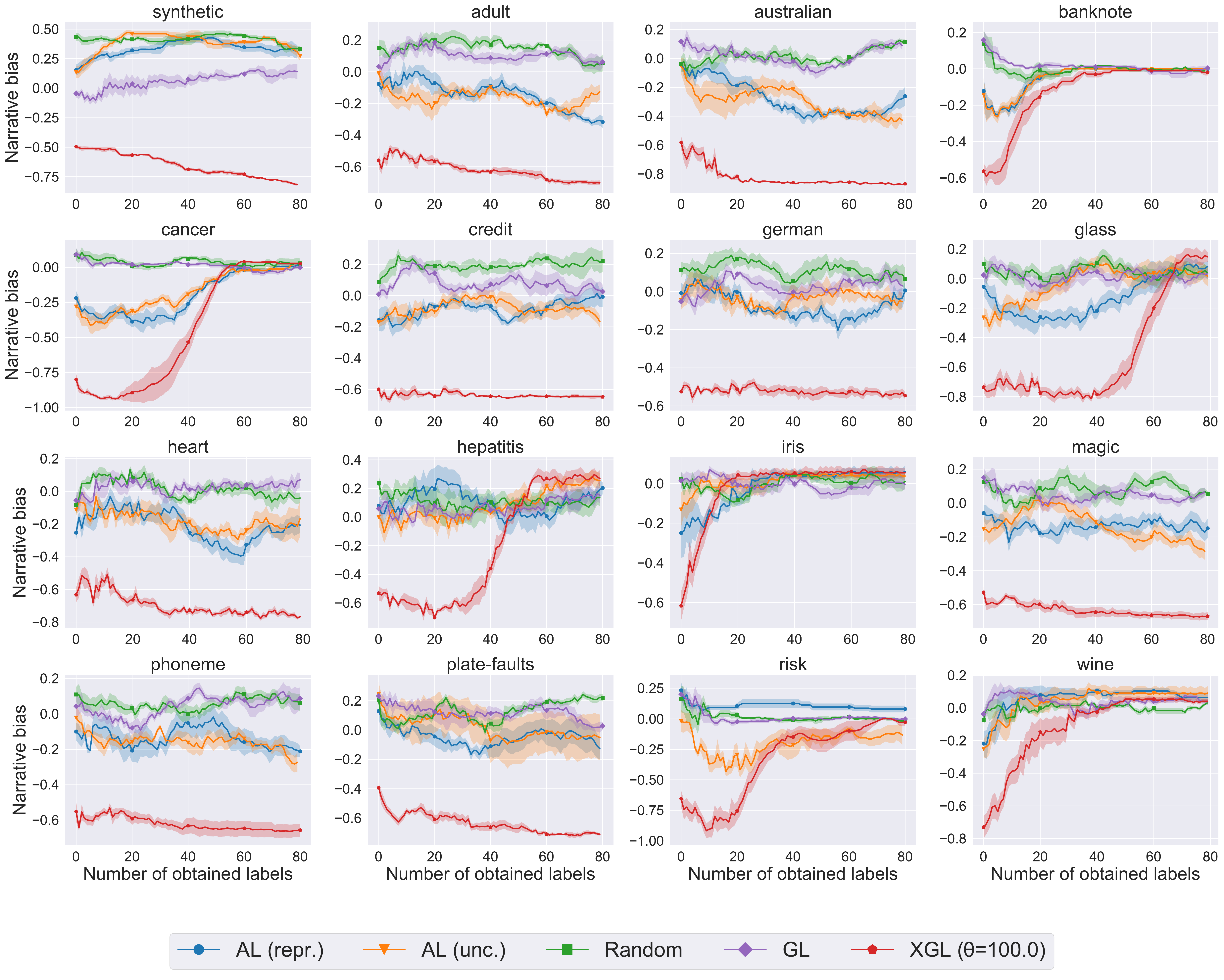}
    \caption{Narrative bias for all methods on the original data sets.}
    \label{fig:narrative_bias_no_extra}
\end{figure*}

\begin{figure*}[p]
    \centering
    \includegraphics[width=\linewidth]{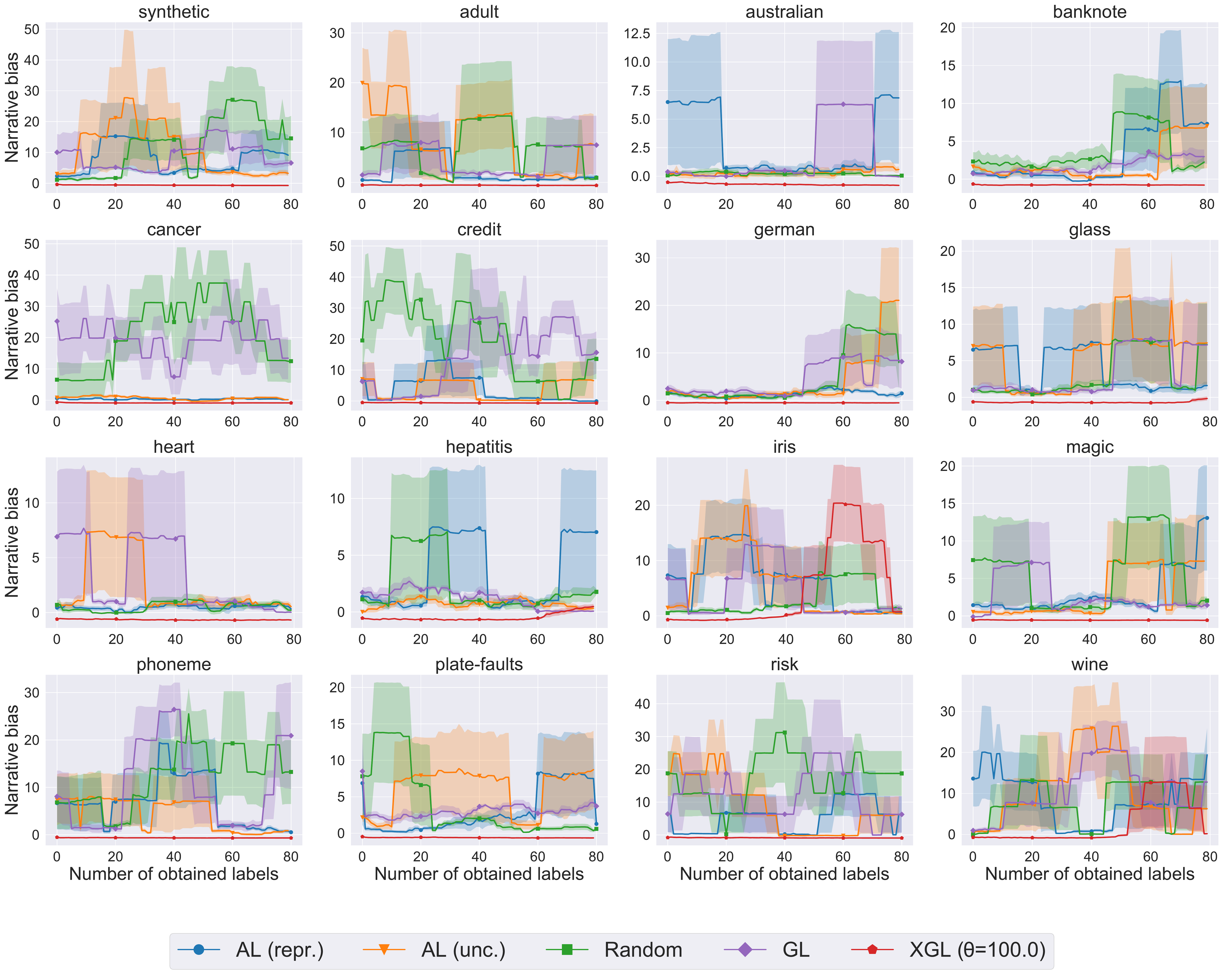}
    \caption{Narrative bias for all methods on the UU-augmented data sets.}
    \label{fig:narrative_bias_weight25}
\end{figure*}

Figures~\ref{fig:narrative_bias_no_extra} and \ref{fig:narrative_bias_weight25} report the narrative bias of XGL and all competitors on the original and ``+uu'' data sets, respectively.  It can be clearly observed that XGL is less affected than all competitors.

A couple of remarks are in order.  The first one is that AL and GL are black-box, and therefore narrative bias would only affect variants which present the user with the prediction associated to the query instances.  The second one is that in XGL the explanation is not query specific, hence the fact that global explanations capture low-performance regions (like unknown unknowns) is reflected in the low NB values shown by the plots for XGL.  Notice that introducing costly UUs (in the ``+uu'' data sets) dramatically increases the narrative bias of all methods except XGL.  The couple of cases in which NB of XGL goes above $0$ are iris and wine, and the condition is only temporary.

\subsection{Predictive performance}

\begin{figure*}[p]
    \centering
    \includegraphics[width=\linewidth]{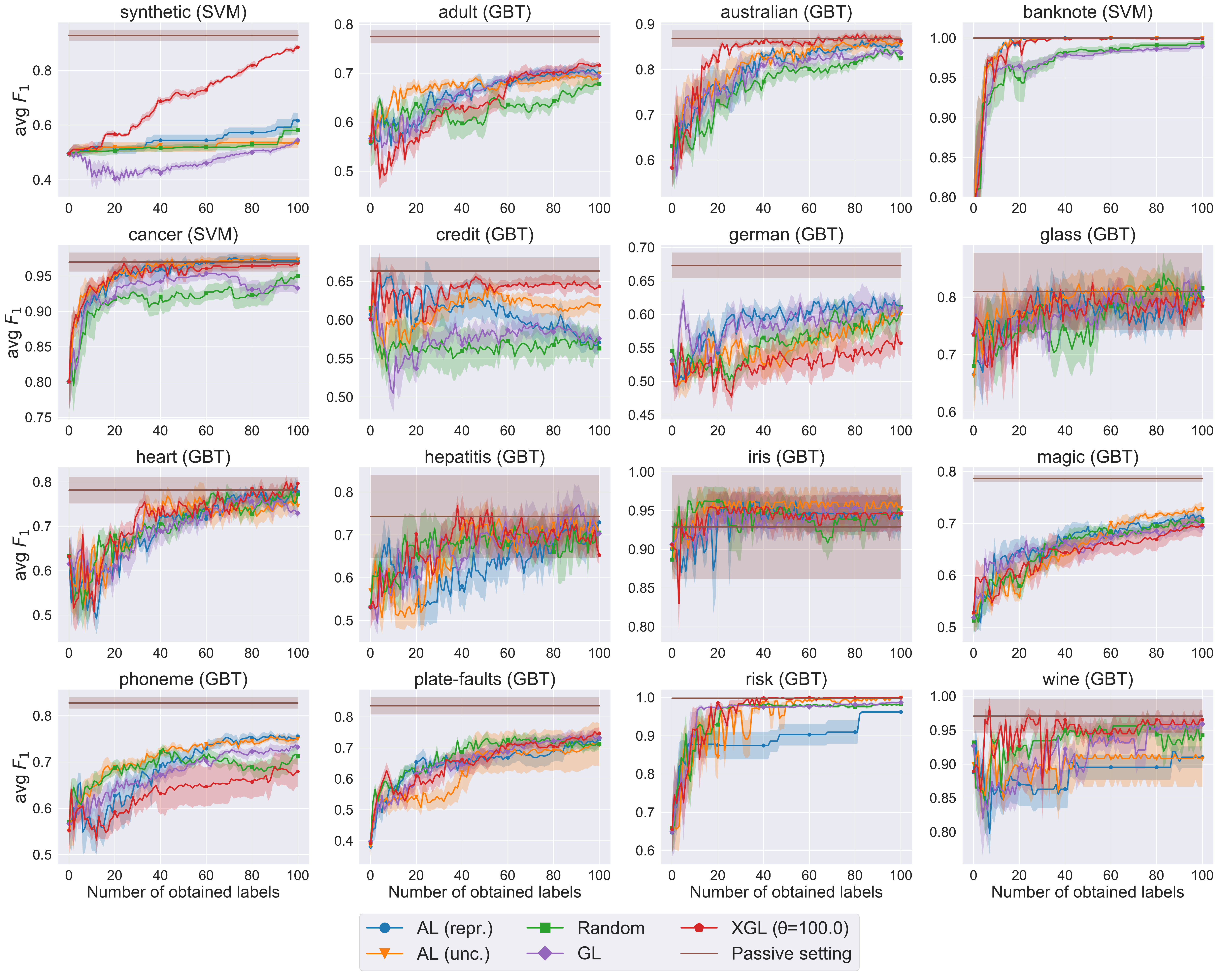}
    \caption{$F_1$ score of \method and the other competitors on all original data sets. }
    \label{fig:results_all_no_extra}
\end{figure*}

\begin{figure*}[p]
    \centering
    \includegraphics[width=\linewidth]{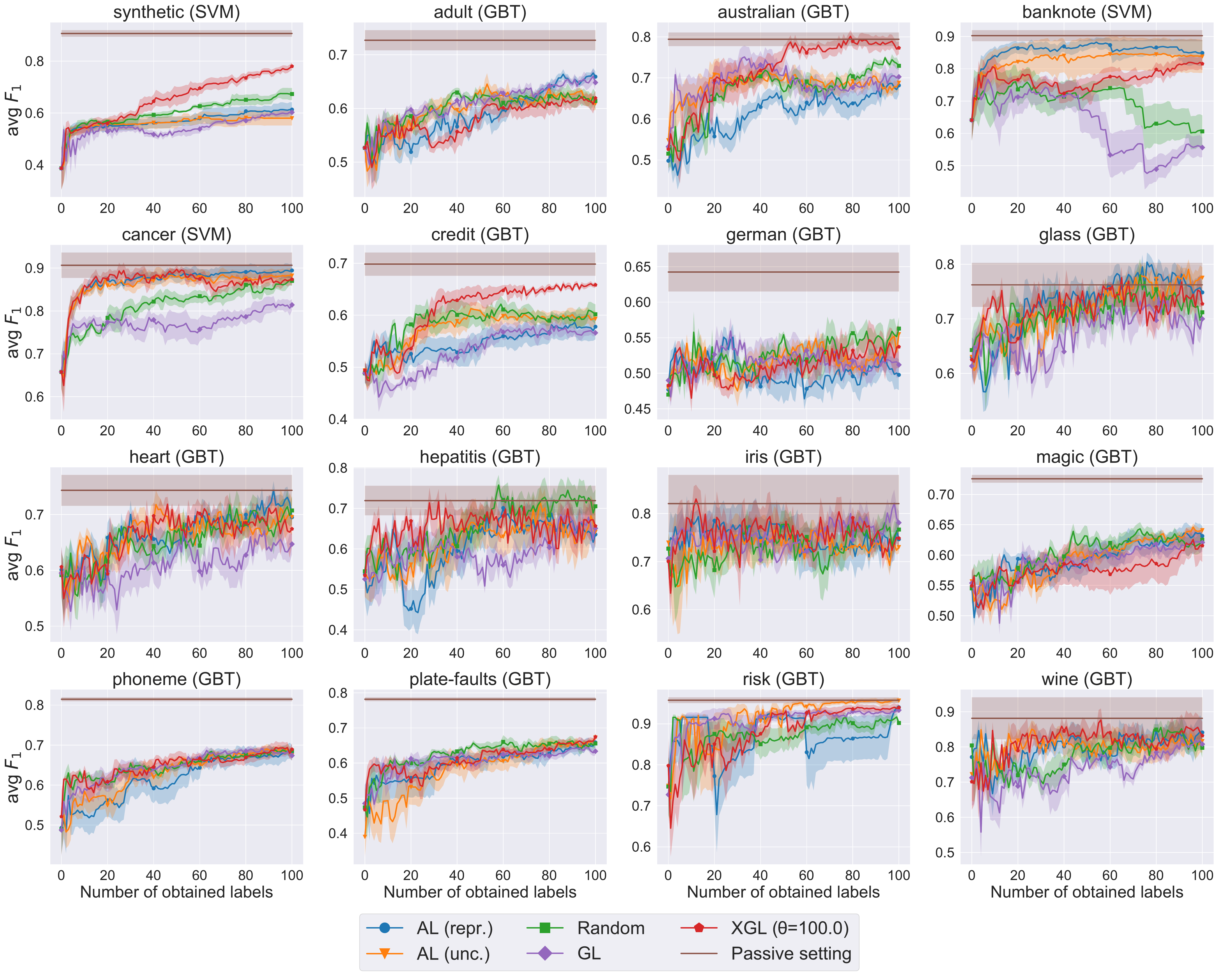}
    \caption{$F_1$ score of \method and the other competitors on the UU-augmented data sets. }
    \label{fig:results_all_weight25}
\end{figure*}

The predictive performance of \method compared to the competitors is shown in Figures~\ref{fig:results_all_no_extra} and \ref{fig:results_all_weight25} for the original and ``+uu'' data sets, respectively. \method typically performs comparably or better than the competitors in most data sets, with the notable exception of adult and magic (and the original phoneme data set).

\subsection{Helpful vs Less Helpful Supervisors}

\begin{figure*}[tb]
    \centering
    \includegraphics[width=\linewidth]{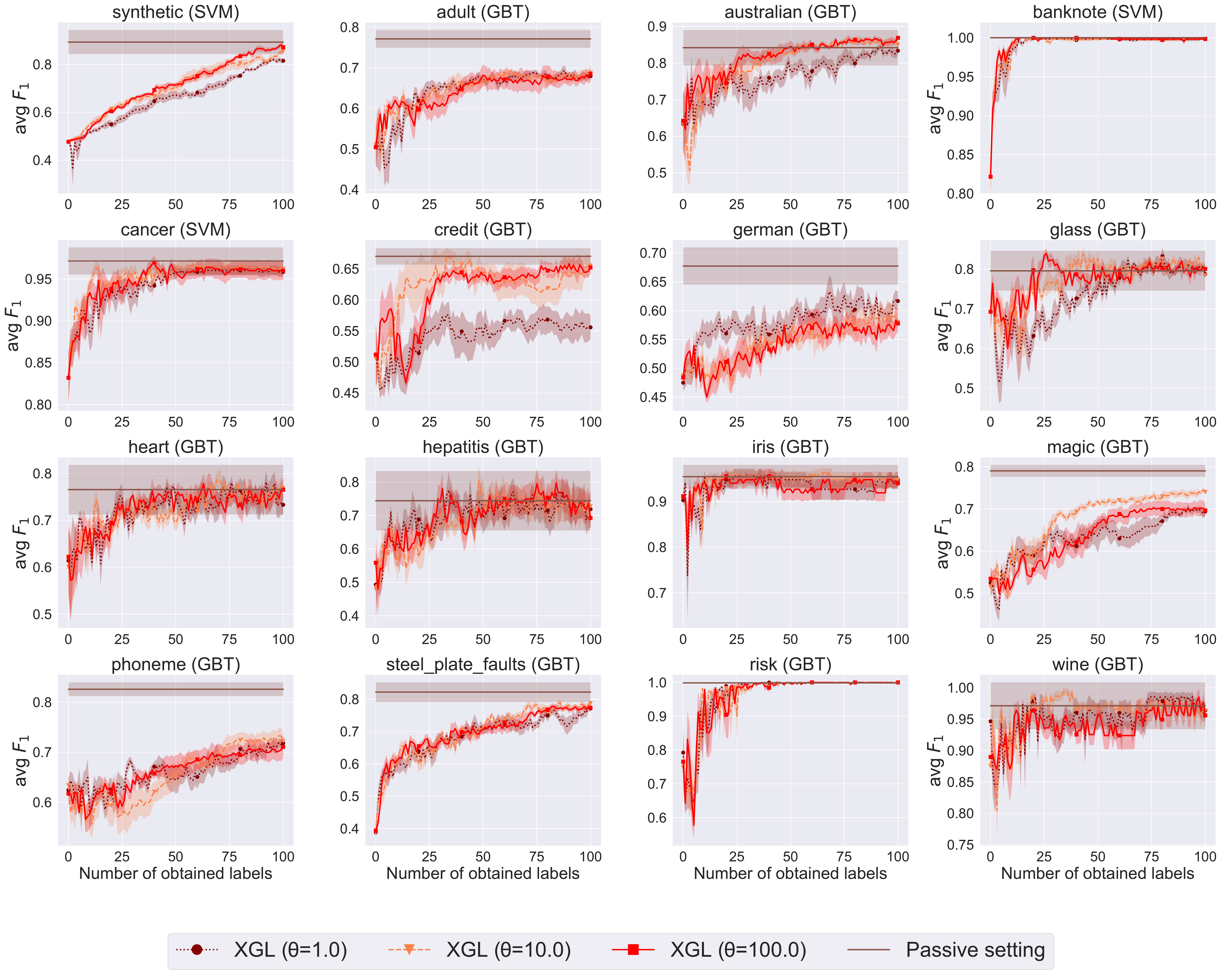}
    \caption{Behavior of \method as the helpfulness $\theta$ of the user simulator changes.}
    \label{fig:rules_thetas}
\end{figure*}

Figure~\ref{fig:rules_thetas} reports the behavior of XGL on the original data sets by varying the values of $\theta = 1, 10, 100$, for simulated users.  Higher the value of theta, more expert is the user.

\end{document}